\documentclass[10pt,journal,compsoc]{./IEEEtran}

\usepackage[cmex10]{amsmath}

\usepackage[nocompress]{cite}
\usepackage{url}

\usepackage{epsfig}
\usepackage{graphicx}
\usepackage{amsmath}
\usepackage{amssymb}
\usepackage{color,eucal,xspace}

\graphicspath{{./}{./figs/}{../}{../figs/}}

\usepackage[caption=false,font=small,labelfont=sf,textfont=sf]{subfig}

\usepackage{fixltx2e}

\begin{document}
\title{Cross-convolutional-layer Pooling for Image Recognition}

\author{Lingqiao Liu, Chunhua Shen, Anton van den Hengel

\IEEEcompsocitemizethanks{\IEEEcompsocthanksitem
The authors are with the School of Computer
Science, The University of Adelaide, Adelaide, SA 5005, Australia.
(E-mail: \{lingqiao.liu, chunhua.shen, anton.vandenhengel\}@adelaide.edu.au).
}
\thanks{}}

\markboth{Accepted to IEEE Transactions on Pattern Analysis and Machine Intelligence, October 2016}%
{Liu \MakeLowercase{\textit{et al.}}: Cross-layer pooling for image recognition}
\IEEEtitleabstractindextext{%

\begin{abstract}

  Recent studies have shown that a Deep Convolutional Neural Network (DCNN)
  trained on a large image dataset can be used as a universal image descriptor
  and that doing so leads to impressive performance for a variety of image
  recognition tasks. Most of these studies adopt activations from a single DCNN
  layer, usually
	a
	fully-connected layer, as the image representation. In
  this paper, we proposed a novel way to extract image representations from two
  consecutive convolutional layers: one layer is used  for local feature
  extraction and the other serves as guidance to pool the extracted features.
  By taking different viewpoints of convolutional layers, we further develop
  two schemes to realize this idea.  The first  directly uses convolutional
  layers from a DCNN. The second applies the pre-trained CNN on densely
  sampled image regions and treats the fully-connected activations of each
  image region as a convolutional layer's feature activations. We then train
  another convolutional layer on top of that as the pooling-guidance
  convolutional layer. By applying our method to three popular visual
  classification tasks,  we find that our first scheme tends to perform better on
  applications which demand strong discrimination on lower-level visual
  patterns while the latter excels in cases that require discrimination on
  category-level patterns. Overall, the proposed method achieves superior
  performance over existing approaches for extracting image representations from a
  DCNN. In addition, we apply cross-layer pooling to the problem of image retrieval and
  propose schemes to reduce the computational cost. Experimental results
  suggest that the proposed method achieves promising results for the
  image retrieval task.

\end{abstract}
}

\maketitle

\tableofcontents
\clearpage

\IEEEdisplaynotcompsoctitleabstractindextext
\section{Introduction}

Recently, Deep Convolutional Neural Networks (DCNNs) have attracted much
research attention in visual recognition, largely due to their excellent
performance \cite{ImageNetDeepLearning}. It has been discovered that the
activation of a DCNN trained on a large dataset, such as ImageNet
\cite{ImageNet}, can be employed as a universal image descriptor, and applying
this descriptor to many visual classification and retrieval problems delivers
impressive performance~\cite{CNN_Baseline,ArxivNewBaseline,
conv_pool_retrieval}. This discovery quickly sparked significant interest and
inspired many extensions, including \cite{CNN_Regional,Our_NIPS}. A fundamental
issue with these kinds of methods is how to generate an image representation
from a pre-trained DCNN. Most current solutions take activations of a single
DCNN layer, usually the fully-connected layer, as the image representation.

In this paper, we show that we can build a powerful image representation using
the activations from two consecutive convolutional layers. We name our method
cross-convolutional layer pooling (or cross-layer pooling for short). This new
method relies on two crucial components: (1) we extract local features from one
convolutional layer (2) we pool extracted local features by using activations
from its successive convolutional layer as guidance.

The first component is motivated by recent work
\cite{CNN_Regional,Our_NIPS,YaoCVPR2015} which has shown that DCNN activations
are not translation invariant and that it is beneficial to extract fully
connected layer activations from a DCNN to describe local regions and create
the image representation by pooling multiple regional DCNN activations. In this
paper, we view those regional CNN activations as a newly added convolutional
layer (named as the augmented convolutional layer as discussed in section
\ref{sect:fc_vs_conv}). Inspired by this view, we also extract local features
from the original convolutional layers of the pre-trained CNN. %

The second component is motivated by the parts-based pooling method
\cite{zhangningpos} which was originally proposed for fine-grained image
classification. This method creates one pooling channel for each detected part
region while the final image representation is obtained by concatenating pooling
results from multiple channels. We generalize this idea to
DCNNs and avoid the need for annotating predefined parts. More specifically, we
deem the feature map of each filter in a convolutional layer as the detection
score map of a part detector and apply the feature map to weight regional
descriptors extracted from the previous convolutional layer in the pooling
process. The final image representation is obtained by concatenating pooling
results from multiple channels with each channel corresponding to one feature
map. Note that in contrast to existing regional-DCNN based methods
\cite{CNN_Regional,Our_NIPS}, the proposed method does not require additional
dictionary learning and encoding steps at either the training or testing stage once
the convolutional layer activations become available. To further reduce the
memory use in storing image representations, we also experiment with a coarse
`feature sign quantization' compression scheme and show that the discriminative
power of the proposed representation can be largely maintained after
compression.

Besides image classification,  we explore the use of cross-layer pooling for image retrieval.
To overcome the high computational cost of the direct implementation of
cross-layer pooling,  we propose to employ feature binarization and adaptive
pooling channel selection schemes to
 reduce the computational cost.

We conduct extensive experiments on three popular visual classification
datasets, and three popular image retrieval datasets. Experimental results
suggest that the proposed method
achieves
significantly better performance
than competitive methods in most cases. Further ablation studies provide insight into the
importance of various components of our approach.

A preliminary version of this paper has been published in \cite{CL_CVPR}. In
this paper, we have made a significant extension. The major differences are
threefold.
\begin{enumerate}
  \item
 We view the scheme of extracting fully connected CNN
activations at densely sampled regions as a newly added convolutional layer and
perform cross-layer pooling at that level. This extension makes our method more
widely applicable.
\item
  We apply cross-layer pooling to image retrieval tasks
and propose new schemes to reduce the computational cost.
\item
  We have conducted more experiments to validate the proposed method,
including experiments with a better CNN model and new ablation studies.
\end{enumerate}

\begin{figure*}[t!]
	\centering
    \includegraphics[height=80mm]{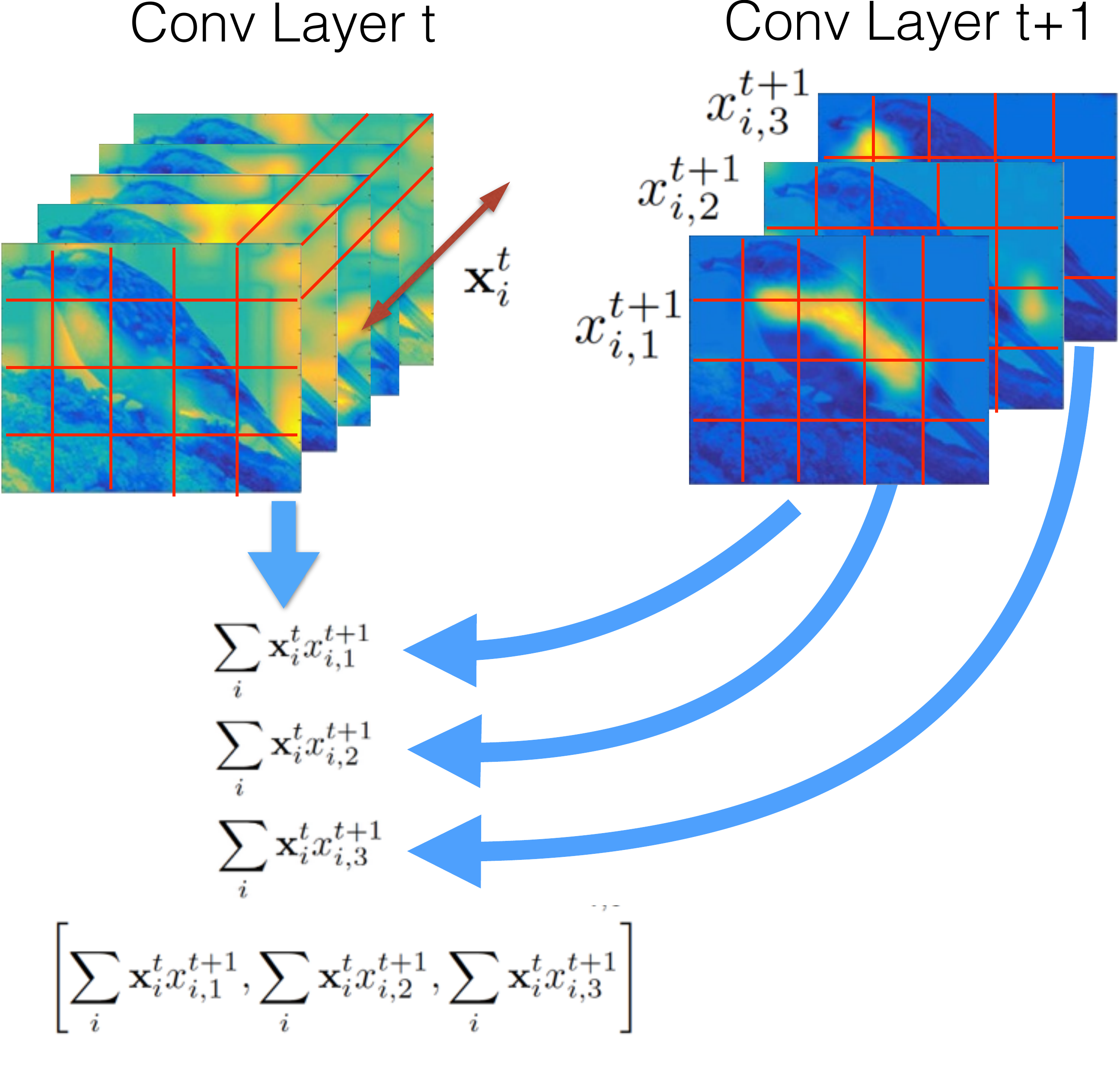}
    \caption{Illustration of the proposed method. Our method performs the pooling operation by using two consecutive convolutional layers. Local features $\mathbf{x}_i^t$ are extracted from the $t$th convolutional layer, and each  of the feature maps at the $(t+1)$th convolutional layer is used to perform weighted sum-pooling for $\mathbf{x}_i^t$. The concatenation of all pooling results is used as the image representation. }
	\label{fig:overview}
\end{figure*}

\noindent
\textbf{Preliminary:}
Two network structures are considered in this paper. One is the AlexNet \cite{ImageNetDeepLearning}
and another one is the VGG very deep (VGGVD in short) network \cite{VGGVD}. Both networks are composed by the cascade of convolutional layers and fully connected layers.
 At each convolutional layer, multiple filters are applied, and it results in multiple feature maps, one for each filter. In this paper, we use the term `feature map' to indicate the convolutional result (after applying the ReLU) of one filter and the term `convolutional layer activations' to indicate feature maps of all filters in a convolutional layer.

\section{Related Work}
\label{sect:existing_ways}
In the literature, there are two primary methods for using a pre-trained DCNN to create an image
representation: (1) directly feeding the whole image into a pre-trained DCNN and extracting its
activations; (2) applying the pre-trained DCNN to subregions of the
input image and aggregating activations from multiple regions as the image representation. Usually,
the first method extracts the last few fully-connected layer activations as the image-level
representation. Fine-tuning is sometimes applied to make the network better adapted to a given task.
Also, to make this kind of method more robust to image transforms, averaging activations from
several jittered versions of the original image, e.g., several slightly shifted versions of the input image, has been employed to obtain better classification performance \cite{ArxivNewBaseline}.

DCNNs can also be applied to extract local features. It has been suggested that DCNN activations are not invariant to a large amount of translation \cite{CNN_Regional} and that performance will be degraded if input images are not well aligned. To handle this issue, it has been suggested to sample multiple regions from an input image and use one DCNN, called regional-DCNN in this scenario, to describe each region. The final image representation is aggregated from activations of those regional-DCNNs \cite{CNN_Regional}. In \cite{CNN_Regional}, another layer of unsupervised encoding is employed to create the image-level representation \cite{CNN_Regional, Our_NIPS}. In \cite{DeepPatternMining}, discriminative patterns are mined from those regional activations for classification. It is shown that for many visual tasks \cite{CNN_Regional,Our_NIPS} this approach leads to better performance than directly extracting DCNN activations as global features.

One common factor in the above methods is that they all use fully-connected
layer activations as features. Some recent studies explore the use of
convolutional layers to extract image representations. For example, the work in
\cite{Deep_texture} applies Fisher vector pooling to the local features
extracted from a convolutional layer to create image representations for
texture classification.   The work in \cite{FastRCNN} uses pooled convolutional
activations for object detection.
The authors in \cite{conv_pool_retrieval}
demonstrated that the pooled convolutional feature are well suited to the image
retrieval task. The work in \cite{BilinearCNN} is probably  most relevant to our work. As
mentioned in \cite{BilinearCNN} itself, their approach is an extension of the
method proposed in~\cite{CL_CVPR} (the preliminary version of this
paper)
for fine-grained image classification. It uses a similar strategy as ours to
combine the convolutional feature activations from two DCNNs and jointly
fine-tune all of the parameters in an end-to-end fashion. \section{Proposed
Method}

\begin{figure}
	\centering
    \includegraphics[height=35mm]{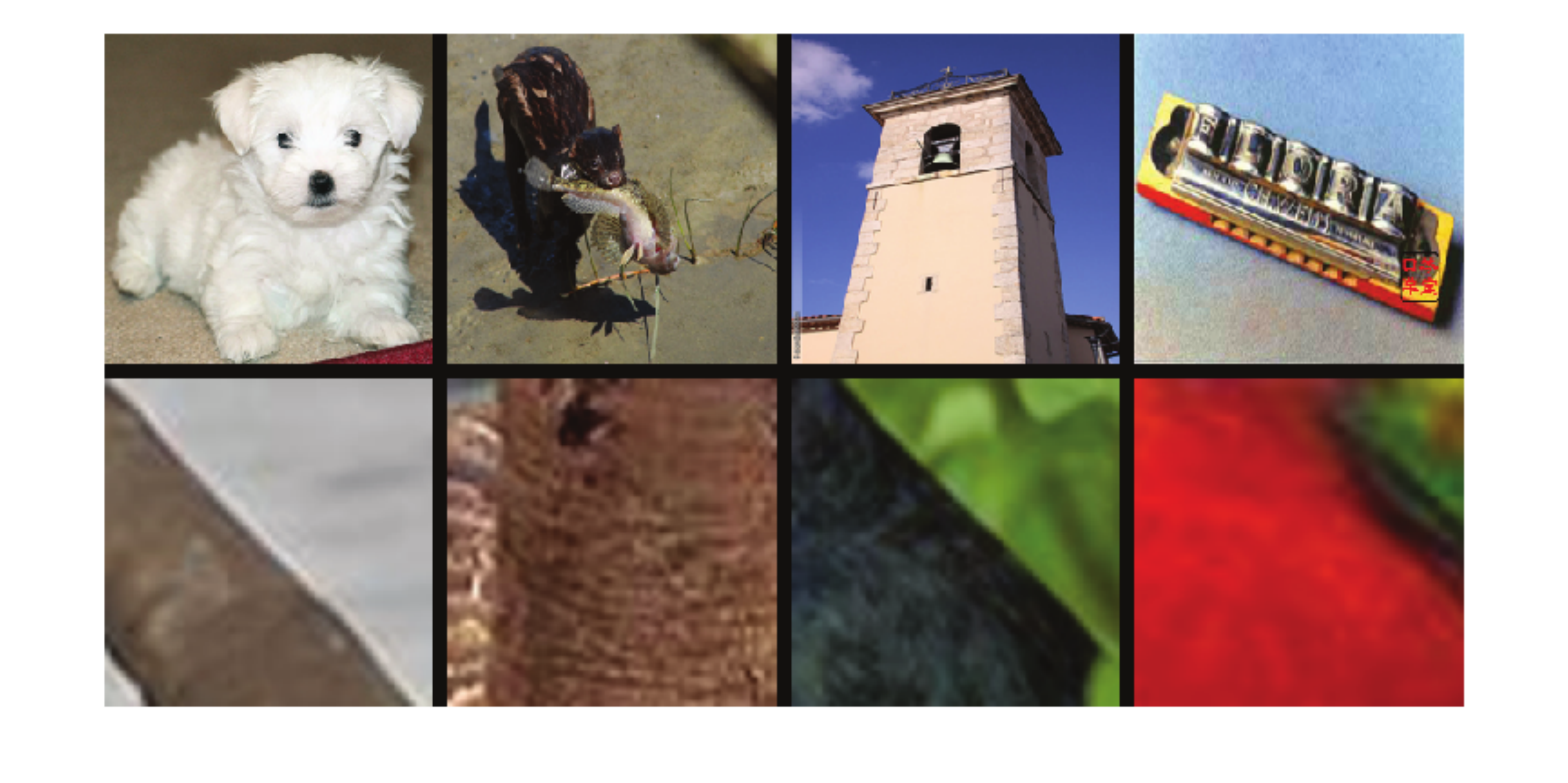}
    \caption{This figure demonstrates the image style mismatch issue when using fully-connected layer activations as regional descriptors.
    Top row: input images that a DCNN `sees' at the training stage. Bottom row: input images that a DCNN `sees' at the test stage.}
	\label{fig:image_vs_region}
\end{figure}

\subsection{Convolutional layers, fully connected layers and notations}\label{sect:fc_vs_conv}

The convolutional layer in a CNN is embedded with rich spatial information. Its activations can be formulated as a tensor of the size $H \times W \times D$, where $H$ and $W$ denote the height and width of each feature map and $D$ denotes the number of feature maps. These activations can alternatively be viewed as an array of $D$-dimensional local features extracted at $H \times W$ locations. In this paper, we denote each of the $H \times W$ locations as a \textbf{spatial unit}, and the $D$-dimensional feature maps corresponding to a location as the \textbf{feature vector
for
a spatial unit}.

The fully-connected layer can be seen as a convolutional layer with the receptive field as the whole image. In recent literature \cite{FC_SpeedUp,YaoCVPR2015,CNN_Regional},
the activations from a fully connected layer are
often used as a descriptor for image regions rather than the whole image. As pointed out in \cite{FullyConvolutional}, if the regions are sampled from the input image over a dense grid, such a descriptor extraction process can also be viewed as applying a convolutional layer. For the sake of clarity, in this paper we refer such a convolutional layer as the ``augmented convolutional layer'' or the AConv layer for short and the convolutional layer of the original pre-trained CNN as the OConv layer.

\subsection{Cross-convolutional-layer Pooling}\label{sect:cl_pooling}
In \cite{CNN_Regional,Our_NIPS }, it has been shown that applying an additional pooling operation on the local features extracted from multiple image regions can significantly boost classification performance. Motivated by these methods, we can design a specific pooling layer and apply it to the local features extracted from a convolutional layer which can be either the OConv layer or the AConv layer.

\begin{figure*}
  \centering
    \begin{tabular}{c}
            \subfloat{ \includegraphics[height=40mm]{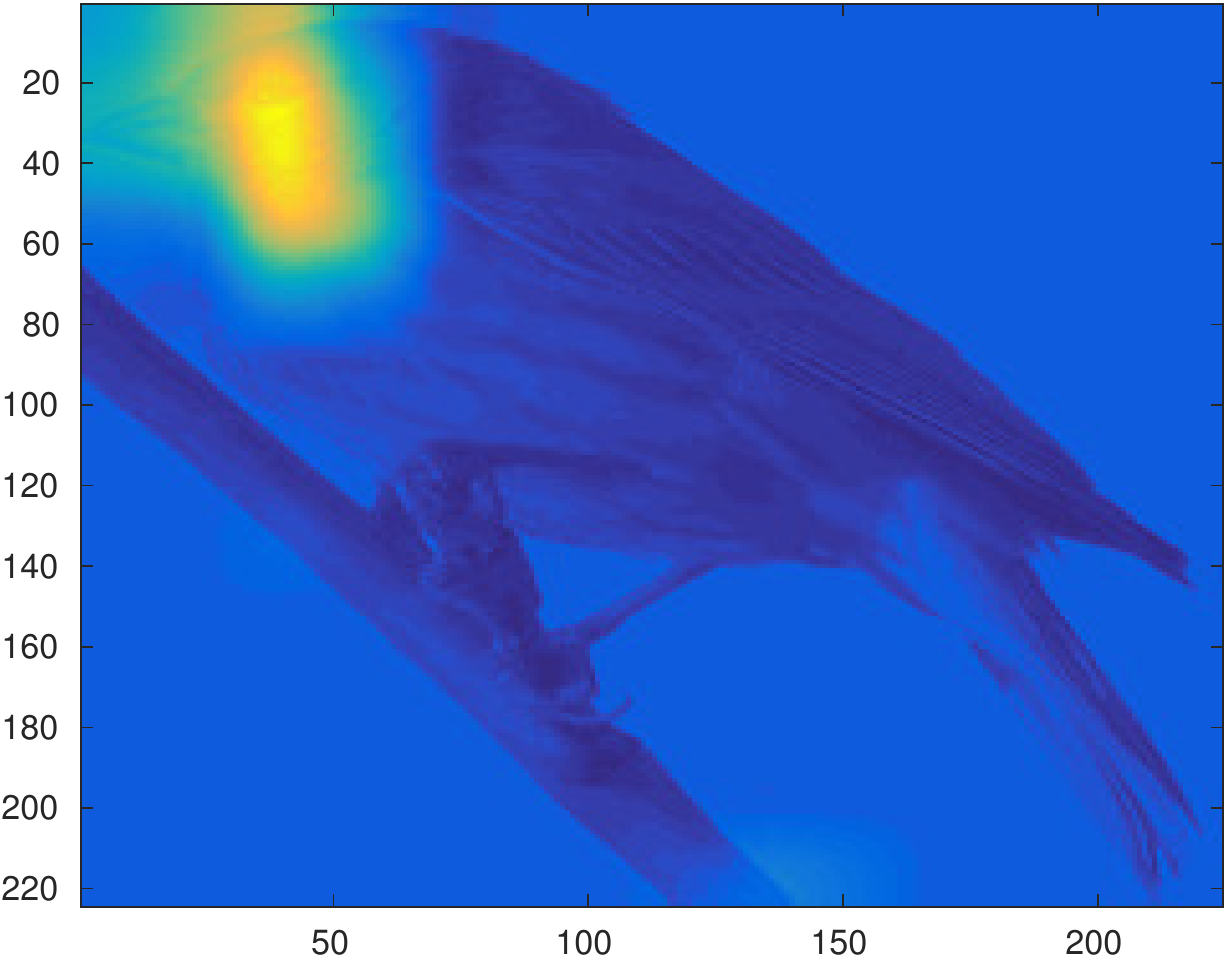}}
            \subfloat{ \includegraphics[height=40mm]{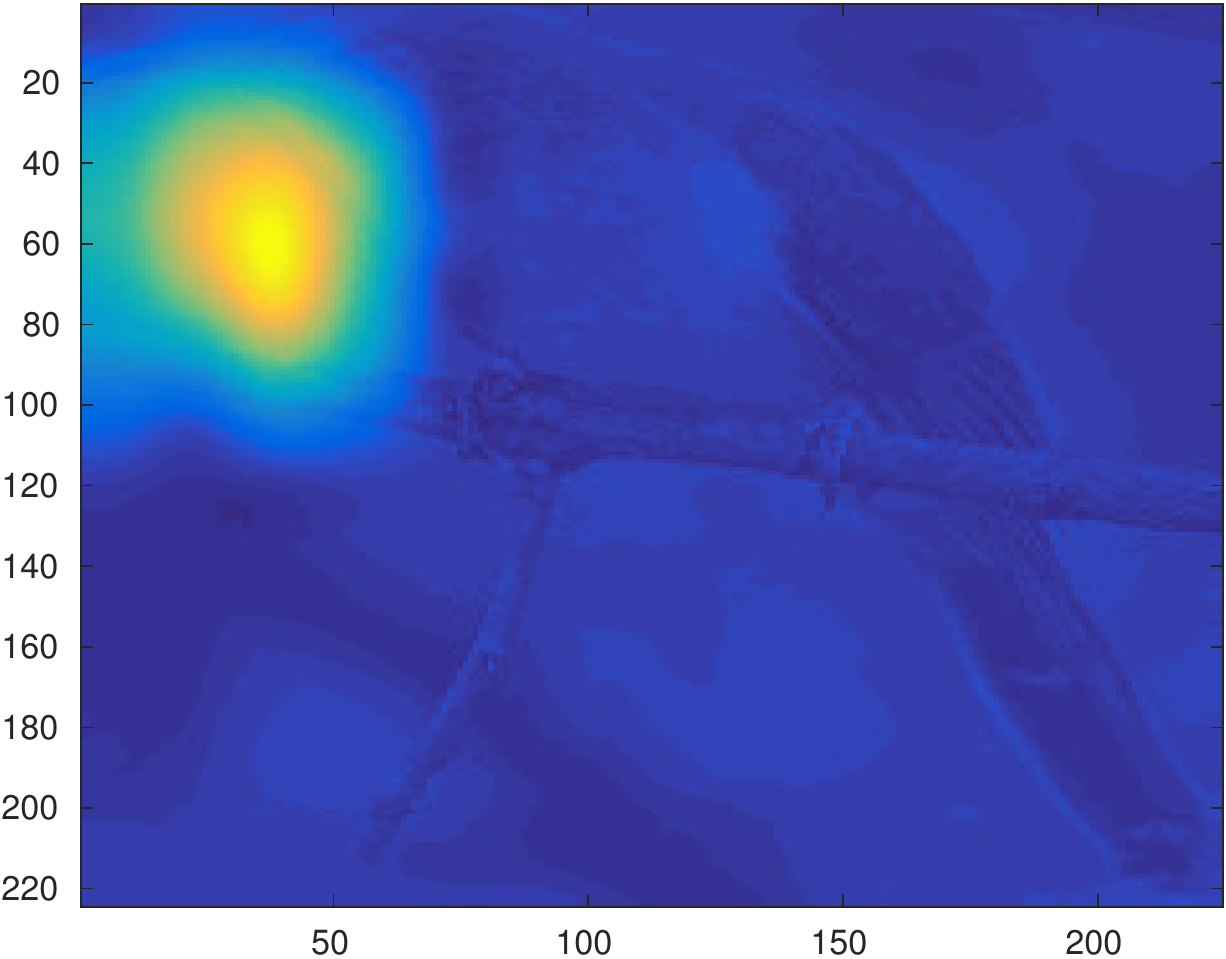}}
            \subfloat{ \includegraphics[height=40mm]{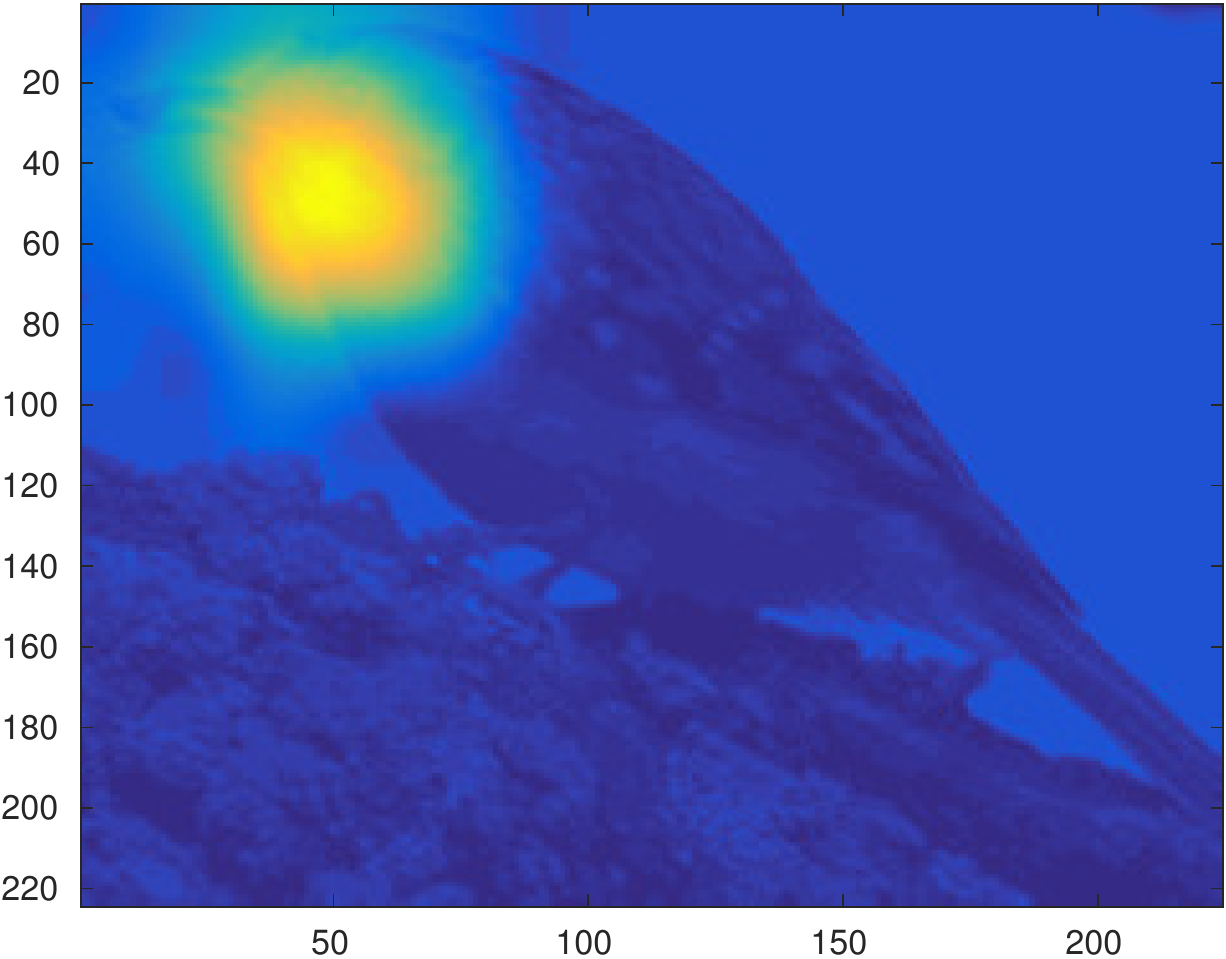}} \\
            \subfloat{ \includegraphics[height=40mm]{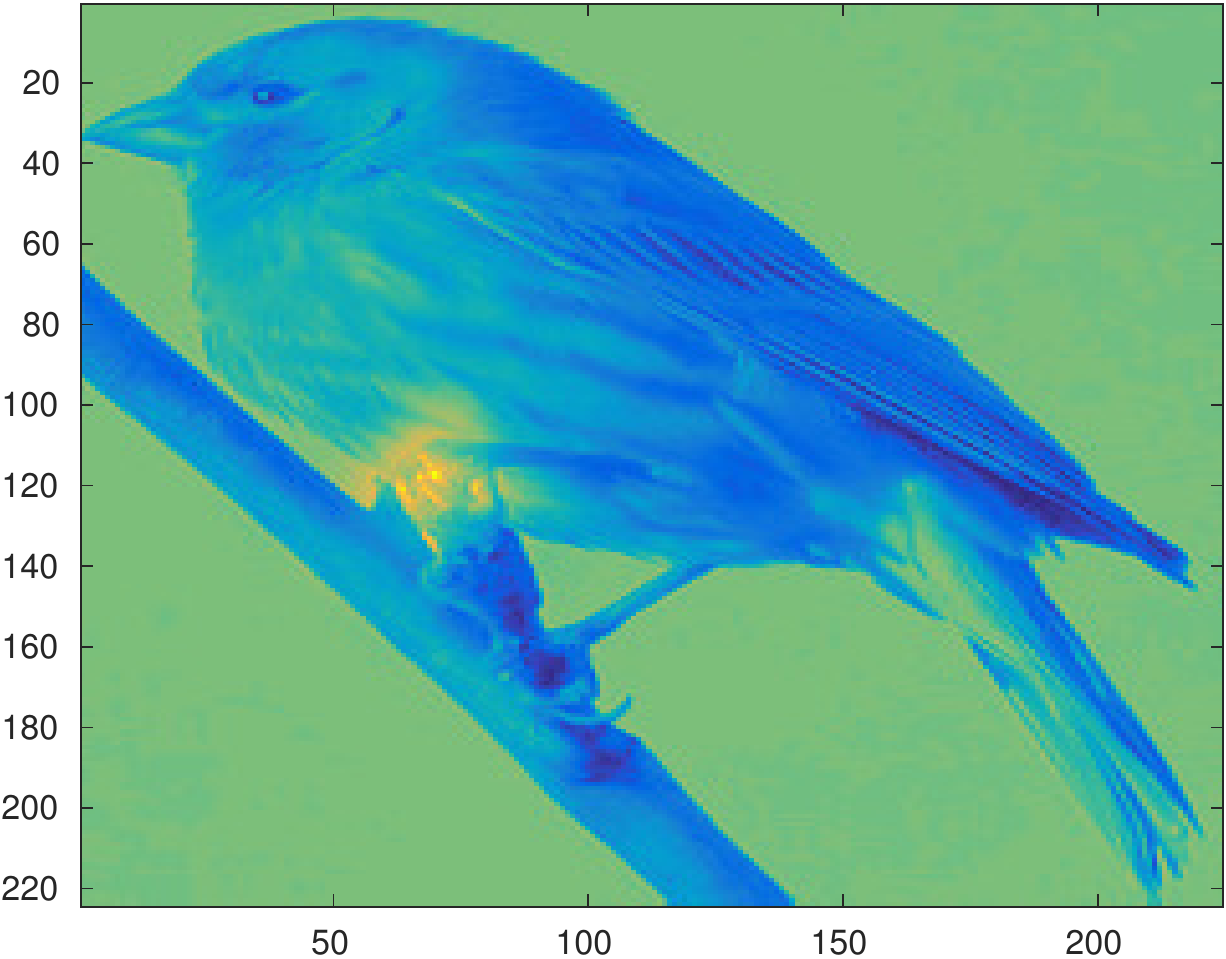}}
            \subfloat{ \includegraphics[height=40mm]{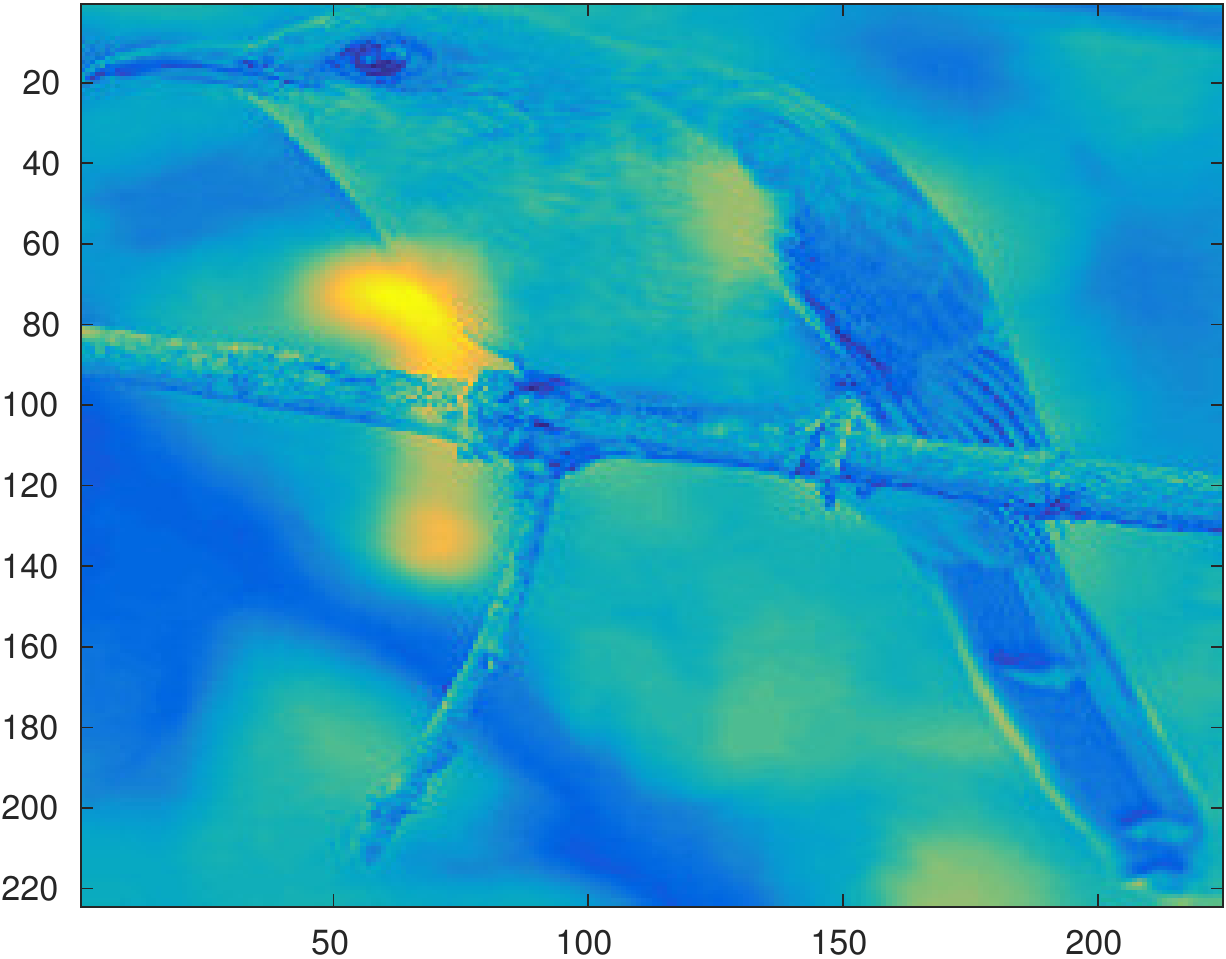}}
            \subfloat{ \includegraphics[height=40mm]{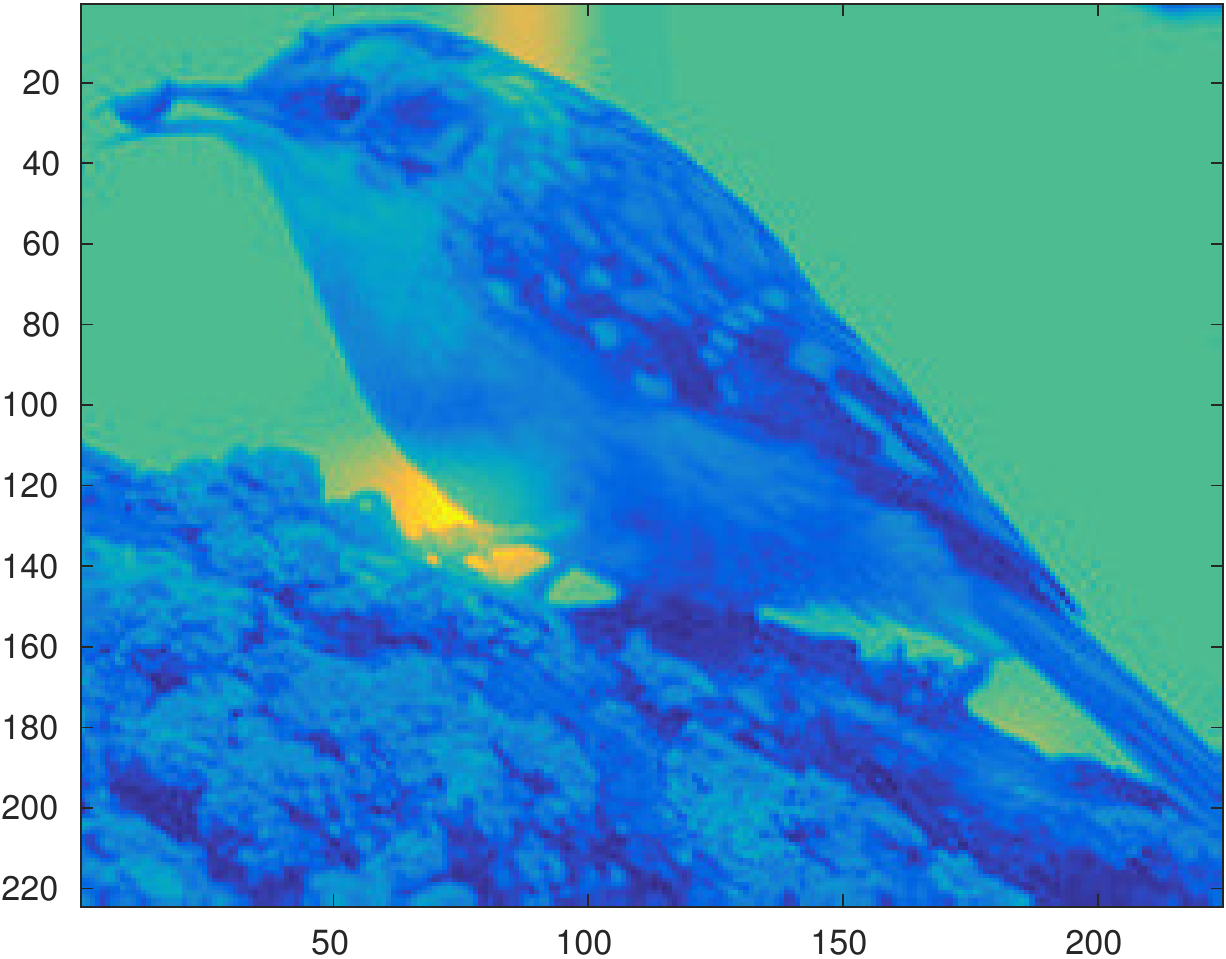}} \\
            \subfloat{ \includegraphics[height=40mm]{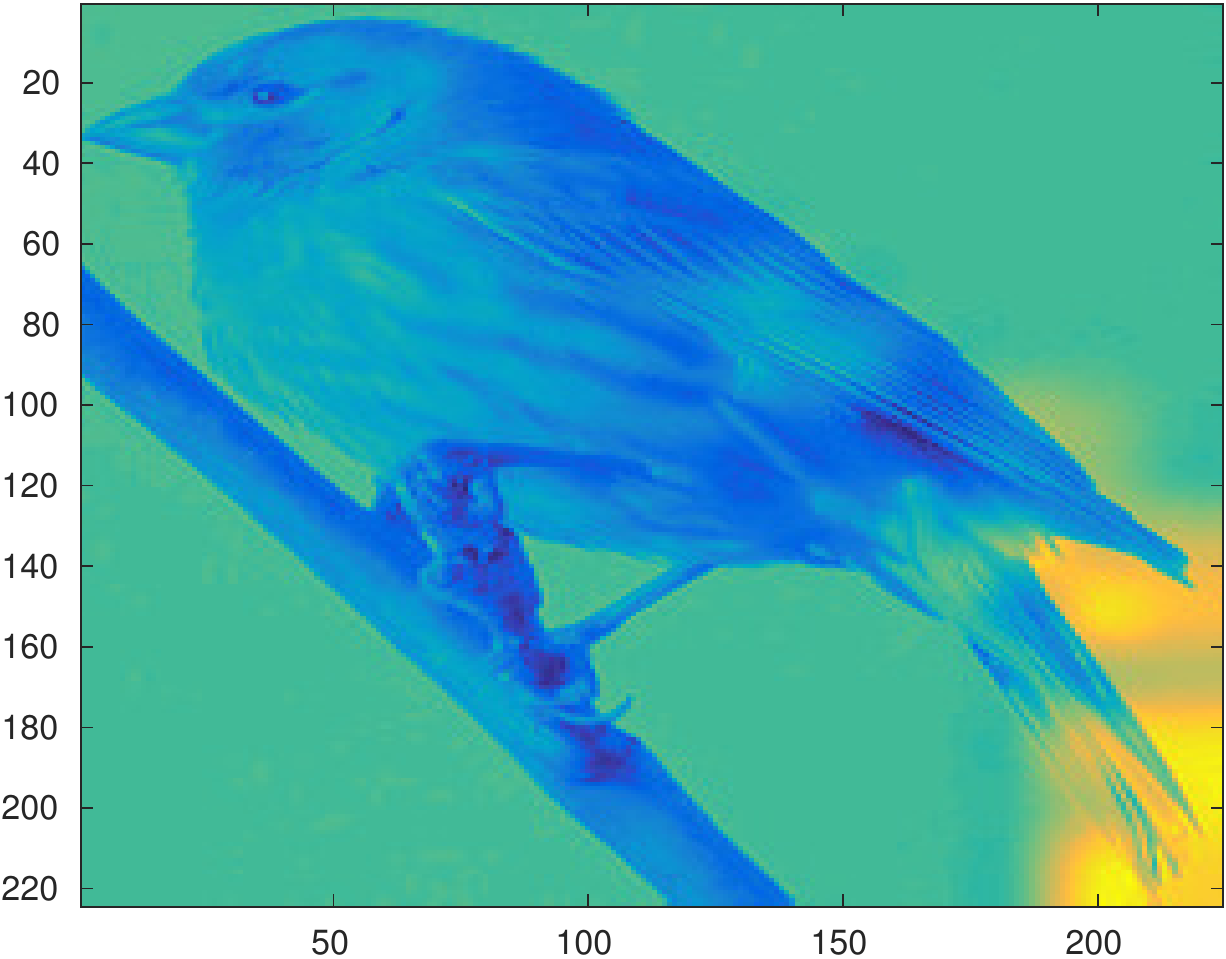}}
            \subfloat{ \includegraphics[height=40mm]{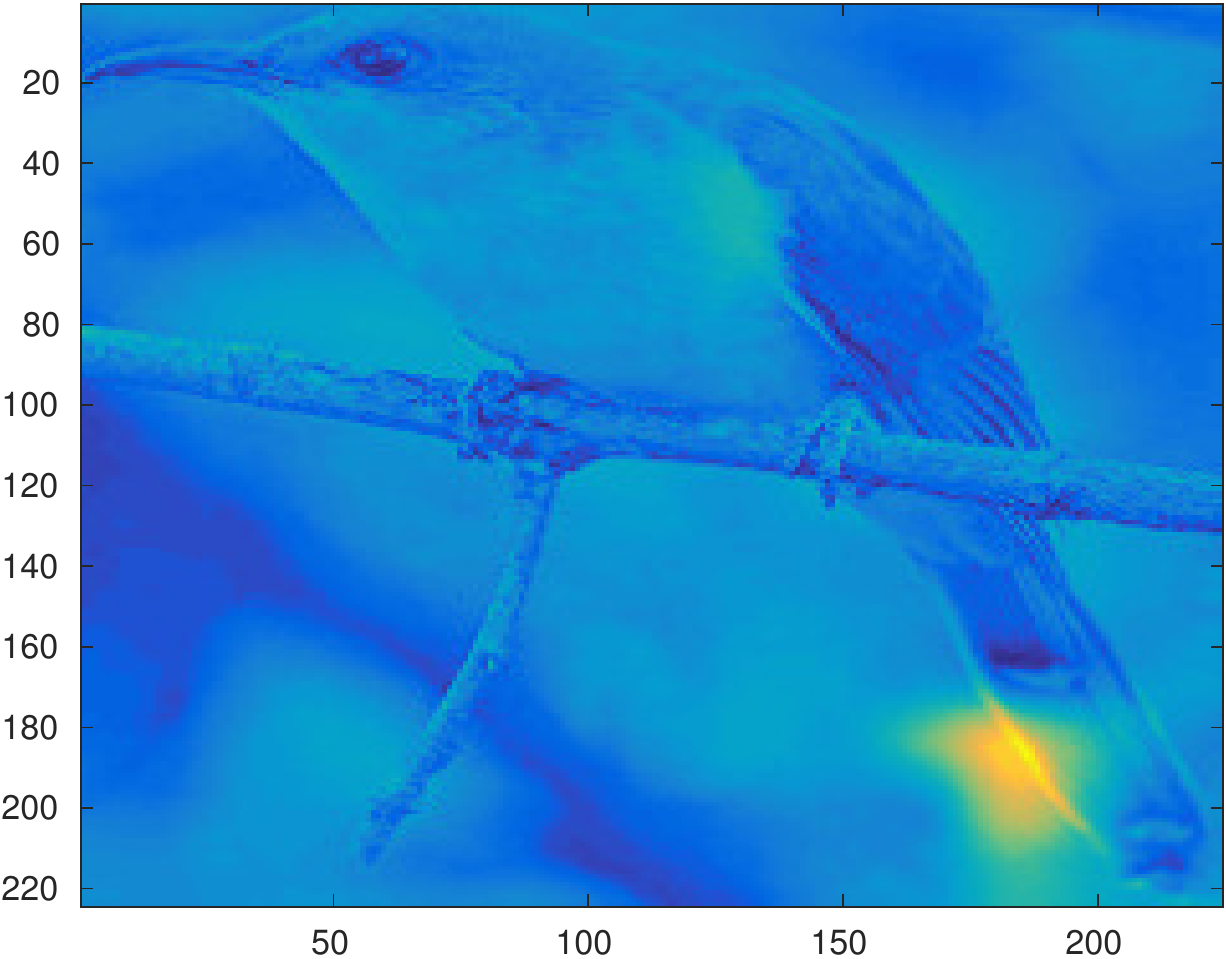}}
            \subfloat{ \includegraphics[height=40mm]{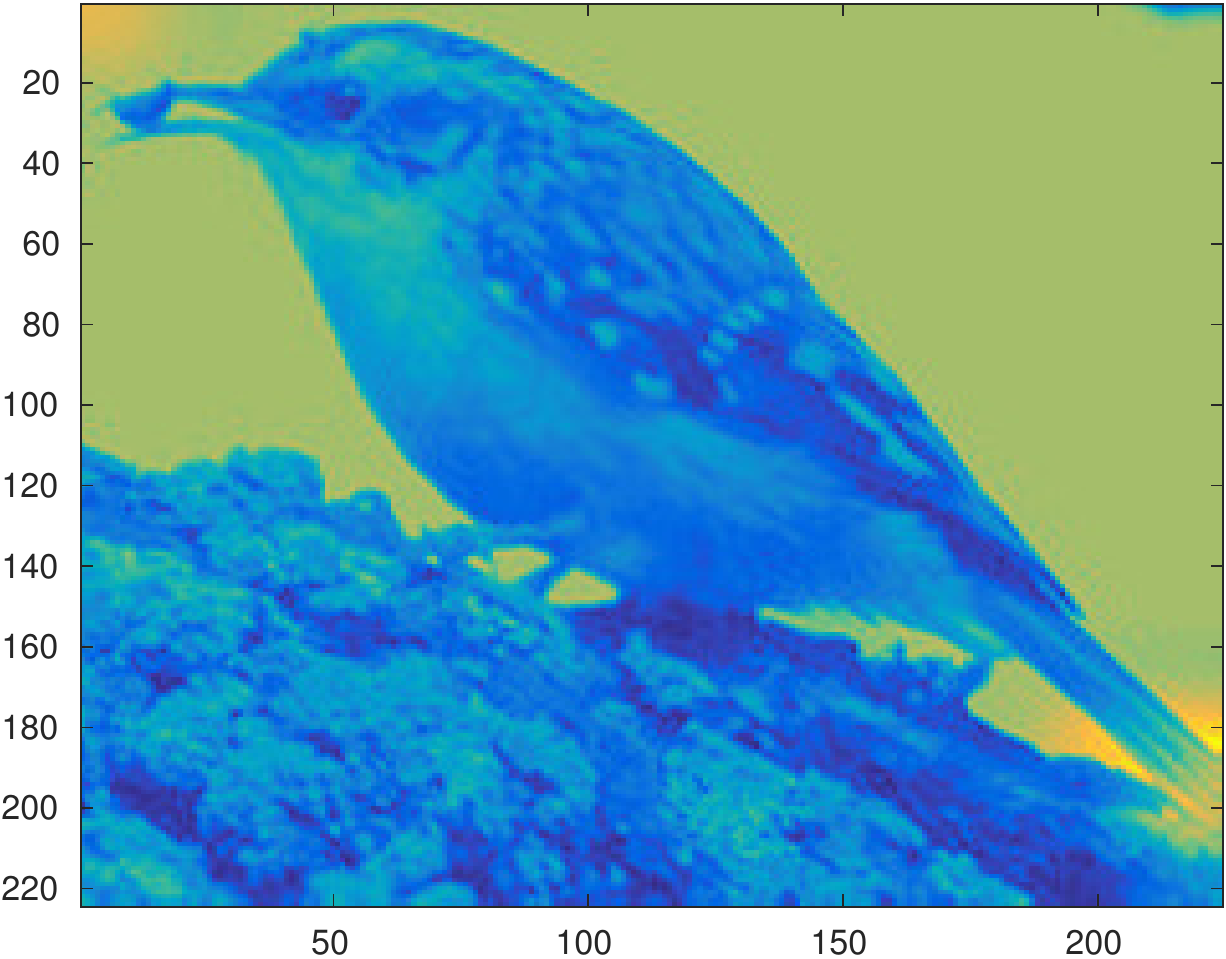}} \\
    \end{tabular}
    \caption{
          Visualization of  feature maps extracted from the conv5-4 layer of the VGG Net. Three feature maps and their activations on three different images are shown. Each row represents the feature map corresponding to the same filter.
          Warmer color indicates higher activation values.
    }
    \label{fig:conv_visualize}
    \end{figure*}

    Various pooling methods could be chosen to aggregate the local features, e.g., max-pooling, sum-pooling or Fisher vector based pooling \cite{ImprovedFV}. In this section, we propose an alternative pooling method which
significantly improves
classification performance.

The proposed method is inspired by the parts-based pooling strategy \cite{zhangningpos} used in
fine-grained image classification. In this strategy, multiple regions-of-interest (ROI) are first
detected, with each corresponding to one human-specified object part, e.g.,
the tails of birds. Then local features falling into each ROI are then pooled together to obtain a pooled feature vector. Given $D$ object parts, this strategy creates $D$ different pooled feature vectors and these vectors are concatenated together to form the image representation. It has been shown that this simple strategy achieves significantly better performance than directly pooling all local features together. Formally, the pooled feature from the $k$th ROI
of the $t$th layer, which we denote by
$\mathbf{P}^{t}_k$, can be calculated by the following equation (let's consider sum-pooling in this case):
\begin{align}\label{Eq:part_pooling}
    \mathbf{P}^{t}_k = \sum_{i = 1} \mathbf{x}_i I_{i,k},
\end{align}
where $\mathbf{x}_i$ denotes the $i$th local feature and $I_{i,k}$ is a binary \textit{indicator map} indicating whether $\mathbf{x}_i$ falls into the $k$th ROI. We can also generalize $I_{i,k}$ to real values with its value indicating the `membership' of a local feature to an ROI. Essentially, each indicator map defines a pooling channel and the image representation is the concatenation of pooling results from multiple channels.

However, in a general image classification task, there are no human-specified part annotations, and even for many fine-grained image classification tasks the annotation and detection of these parts are typically non-trivial. To handle this situation, in this paper, we propose to use feature maps of the $(t+1)$th convolutional layer as $D_{t+1}$ indicator maps. By doing so, $D_{t+1}$ pooling channels are created for the local features extracted from the $t$th convolutional layer. We call this method cross-convolutional-layer pooling or cross-layer pooling for short.

\textcolor{black}{
The use of feature maps as indicator maps is motivated by the observation that a feature map of a
deep convolutional layer is usually sparse and tends to be selective of higher-level visual
concepts, as has also been observed in \cite{VisualizeCNN}. This observation is illustrated in Figure
\ref{fig:conv_visualize}. In Figure \ref{fig:conv_visualize}, we choose three images taken from the
Birds-200 \cite{Birds200} dataset. We sample three feature maps from 256 feature maps in conv5 and
overlay them on the original images for better visualization. As can be seen from Figure
\ref{fig:conv_visualize}, the activated regions of the sampled feature map (highlighted in warm
colors) are semantically meaningful. For example, the activated region in the first row tends to
localize at the head region of a bird while the feature map shown in the second row exhibits high
values around the claw area. Thus, the filter of a convolutional layer works as a part detector, and
its feature map serves a similar role as the part region indicator map. Certainly, compared with the
parts detector learned from human-specified part annotations, the filter of a convolutional layer is
usually not directly task-relevant. However, the discriminative power of our image representation
can benefit from combining a much larger number of indicator maps, e.g.,  256 as opposed to 20-30 (the number of parts usually defined by human).%
\\
Formally, the image representation extracted from cross-layer pooling can be expressed as follows:
\begin{align}\label{Eq:cl_pooling}
	& \mathbf{P}^{t} = [{\mathbf{P}^{t}_1}^\top,{\mathbf{P}^{t}_2}^\top,\cdots,{\mathbf{P}^{t}_k}^\top,\cdots,{\mathbf{P}^{t}}^\top_{D_{t+1}}]^\top \nonumber \\
	& \mathrm{where,~~} \mathbf{P}^{t}_k = \sum_{i = 1}^{N_t} \mathbf{x}^{t}_i x^{t+1}_{i,k}
\end{align}
where $\mathbf{x}_i^t \in \mathbb{R}^{D_t}$ is the $i$th local feature (filter responses) in the $t$th convolutional layer. $N^t$ is the total number of local features at the $t$th layer. $x^{t+1}_{i,k} \in R$ is the activation value of the $i$th spatial unit and the $k$th filter at the $(t+1)$th convolutional layer. Thus $\mathbf{P}^{t}_k$ represents a weighted sum-pooling of $\mathbf{x}_i^t$ with the weight defined by $x^{t+1}_{i,k}$. In total, there are $D^{t+1}$ sets of weights since there are $D^{t+1}$ filters at the $(t+1)$th convolutional layer. The final image representation $\mathbf{P}^{t}$ will be the concatenation of all $\mathbf{P}^{t}_k, k = 1,\cdots,D_{t+1}$ and thus its dimensionality will be $D^{t+1}D^{t}$. Note that here we assume that there is a correspondence between the $i$th local feature at the $t$th layer $\mathbf{x}^{t}_i$ and the $i$th feature activations at the $(t+1)$th layer. This correspondence can be easily established if the consecutive convolutional layers have the same spatial unit layout. For example, the last two convolutional layers in the Alex net both have a 13$\times$13 spatial unit layout and we can deem that feature maps at the same spatial unit across two layers are corresponding. }

Another way of interpreting Equ.~\eqref{Eq:cl_pooling}) is that the image representation is a sum over outer products of corresponding features in two layers. This operation is similar to calculating Gram matrices which have been applied in computer vision \cite{Gatys2015b,Texture_Synthesis2,Texture_Synthesis3}. The difference is that our cross-layer pooling calculates the outer product across different layers and thus can be seen as an extension of the Gram matrices based representation.

\subsection{Augmented convolutional layers vs.\ original convolutional layers}
\label{sect:two_layer_types}

To implement cross-layer pooling, one needs to specify two convolutional layers. In practice, these two convolutional layers can either be chosen from the AConv layer or the OConv layer \footnote{It is also possible to choose one OConv layer and one AConv layer to perform cross-layer pooling. We discuss this possibility in section \ref{sect: different layers}. }. But which type of convolutional layers performs better?
We show empirically below that the best performing layer type depends on the recognition task to which it will be applied.

For the AConv layer, its convolutional filters are the fully-connected layer parameters of the
original CNN. So the AConv layer encodes the higher-level visual concept, e.g.,
object-category-level information.  Thus, if the target problem involves identification of
object-category-level patterns, e.g., to classify whether a ``car'' occurs in the image, then the AConv layer should be used, and its activations can be seen as being similar to object bank detectors \cite{ObjectBank}. Note that even if the problem does not directly involve the identification of an object that appears in the network training task, e.g.,
the 1000 categories in the ImageNet subset, category-level pattern detection may be still beneficial. For example, for scene classification, the occurrence of an object, such as a bed,
can be a strong indicator of the scene class ``bedroom''.

\textcolor{black}{
Compared with the AConv layer, the OConv layer captures lower-level visual patterns. Thus for
target applications which require strong discriminative power to identify lower-level visual
patterns, e.g., specific textures in a fine-grained image classification problem, using the OConv layer can transfer across domain more easily and lead to better performance than using the AConv layer. It should be noted that most commonly used pre-trained CNNs are trained on image classification datasets such as ImageNet \cite{ImageNet}. Thus, if we use an AConv layer to describe low-level patterns, the input images of those pre-trained networks will be very different to that in the target domain. Figure \ref{fig:image_vs_region} shows an example in this case. The top row of Figure \ref{fig:image_vs_region} shows some images from the ImageNet \cite{ImageNet} dataset while the bottom row shows some regions corresponding to the low level visual patterns from the images in Birds-200 \cite{Birds200} dataset. As can be seen, the appearance and the level of detail are quite different between the two rows. Thus, applying the AConv layer to describe lower-level visual patterns will introduce a significant input image style mismatch which could potentially undermine the discriminative power of DCNN activations.}

\subsection{Implementation details}\label{sect:implementation_details}

\noindent \textbf{PCA}: In our implementation, we perform PCA on $\mathbf{x}^{t}_i$ to reduce the dimensionality of $\mathbf{P}^{t}$ to 2000 dimensions for the AConv layer. For the OConv layer, we still perform PCA to de-correlate the local features but without performing dimensionality reduction. We empirically find that this  leads to slightly better performance than using the uncorrected OConv local features.

\noindent \textbf{Normalization}: Since the number of activated spatial units at the guidance convolutional layer can be different for different pooling channels. The pooling vector derived from different channels may have a different energy. Thus, in our implementation we $\ell_2$ normalize the pooled coding vector for each channel. After that, we apply power normalization to $\mathbf{P}^{t}$, that is, we use $\mathbf{\hat{P}}^{t} = \mathrm{sign}(\mathbf{P}^{t})\sqrt{|\mathbf{P}^{t}|}$ as the image representation to further improve performance.

\noindent \textbf{Feature sign quantization}: Besides the aforementioned image representation, we also tried directly using $\mathrm{sign}(\mathbf{P}^{t})$ as an image representation, that is, we coarsely quantize $\mathbf{P}^{t}$ into $\{-1,1,0\}$ according to the feature sign of $\mathbf{P}^{t}$. %

\noindent \textbf{Adding a new convolutional layer for the AConv layer}: One issue when using the
AConv layer for cross-layer pooling is that we need to find two consecutive AConv layers. These two
layers can be obtained by using two consecutive fully-connected layers in the original DCNN.
However, since the fully-connected layers in most commonly used DCNN models have a very large number
of output neurons, e.g., 4096 or 1000. Directly performing cross-layer pooling on those AConv layers
will result in a very high dimensional image representation. To solve this issue, we only utilize
one fully connected layer from the original DCNN as one AConv layer, and stack another newly added
convolutional layer on top of it with a much smaller number of filters, e.g., 100. Then we train the new convolutional layer on the target dataset. The network architecture of our implementation is as follows: a max-pooling layer is applied to pool the activations of the newly added convolutional layer and the pooled result is feed into a logistic regression layer. The negative entropy loss is then utilized to train the new convolutional layer.

\subsection{Application to image retrieval}
Recently, it has been discovered that the pooled convolutional layer activations can form a good image representation for image retrieval \cite{conv_pool_retrieval}. Inspired by the success of the work in \cite{conv_pool_retrieval}, we apply our cross-layer pooling method to image retrieval. Since cross-layer pooling creates multiple pooling channels with each pooling channel capturing one type of visual pattern, and the pooling result of each channel is normalized, an image representation created by cross-layer pooling can depict various aspects of visual patterns within the image in a balanced way. In comparison, the representation generated by the direct pooling method in \cite{conv_pool_retrieval} may be dominated by patterns that occur more frequently within the image.

Directly applying cross-layer pooling for image retrieval will incur a high computational cost due to the high dimensionality of the generated image representations. For example, if there are $M$ pooling channels, the computational cost can be $M$ times higher than the method in \cite{conv_pool_retrieval}. To handle this drawback, in this paper we propose two strategies. The first strategy is to binarize the cross-layer pooling feature. This is inspired by the observation (which will be experimentally demonstrated in Section~\ref{sect:retrieval experiment}) that keeping the sign of feature values \footnote{As will be discussed in the second strategy, some channels will be discarded during feature similarity calculation, which is equivalent to setting the feature values within the discarded channels to 0. In this view, there are still three possible values, 1,$-1$,0, in the resulting image representation.
}
does not significantly impact the
discriminative power of cross-layer pooling. The second strategy is to adaptively select a small number, say $k$, of pooling channels for \textit{each query image} and then only retain the features pooled from the selected channels in both the query and the reference image to perform the similarity comparison. Formally, for such a scheme the image similarity between a query image and a reference image is calculated via:
\begin{align}\label{Eq:CL_Retrieval}
    s(\mathbf{I}_q,\mathbf{I}) = \sum_{k \in \mathcal{S}} \mathbf{P}_{q,k}^T \mathbf{P}_{x,k},
\end{align}
where $\mathbf{I}_q$ and $\mathbf{I}$ are the query image and a reference image. In the original
cross-layer pooling, both the query image and the reference image are represented by $D$ subvectors
which are pooled from each pooling channel. We use $\mathbf{P}_{q,k}$ and $\mathbf{P}_{x,k}$ denote
the $k$th subvectors of the query and a reference image respectively.
For the original cross-layer
pooling approach, the comparison should be made over all $D$ subvectors. Here in Equ.~(\ref{Eq:CL_Retrieval}), only subvectors whose indices fall within a small subset $\mathcal{S}$
($|\mathcal{S}|= k \ll D$) are compared.
Thus, the computational cost can be greatly reduced in
comparison with the naive implementation of cross-layer pooling for image retrieval. In this paper,
we construct the set $\mathcal{S}$ by selecting channels (feature maps of a convolutional layer)
with top $k$ average activations. By applying this criterion, the convolutional feature maps with
less significant activations will be discarded. Thus, for this operation, one additional benefit
besides a reduction in the computational cost is that it might suppress the noise patterns and
therefore potentially improve retrieval performance. Note that besides selecting channels with top
$k$ activation values, other criteria can be applied. For example, if the retrieval task is to find
a specific object type such as
sculpture, cloth, an importance weight for each pooling channel can be learned by using images which contain the object-of-interest as positive training samples and random images as negative training samples.

\section{Experiments}

We have organized our experiments into three parts. The first  evaluates the proposed cross-layer pooling method for the image classification application. In the second part, we conduct ablation studies and demonstrate the impact of the various components of our method. In the third part, we evaluate the proposed method on image retrieval tasks. The focus of the third part is to evaluate whether the proposed cross-layer pooling leads to better performance than the method in \cite{conv_pool_retrieval} which also uses a convolutional layer pooling strategy for image retrieval.

\subsection{Image classification experiments}

\subsubsection{Experimental protocol}

\begin{table}[h]
          \caption{Comparison of results on MIT-67. The lower part of this table lists some results reported in the literature.  }
		\centering
		\label{table:MIT67_Result}
	\scalebox{.952}
{
    \begin{tabular}{llll}
            \hline\noalign{\smallskip}
                Methods  &    Accuracy  & Remark/Setting \\
            \noalign{\smallskip}
            \hline
            \noalign{\smallskip}
			 CNN-Global				& 57.9\%          &   AlexNet               \\
			 CNN-Jitter				& 61.1\%            &   AlexNet             \\
			 SCFV	\cite{Our_NIPS} 			& 59.2\% 	        &   AlexNet, OConv   \\
			 CrossLayer (proposed)        &  63.0\%           &   AlexNet, OConv      \\
			 SCFV	\cite{Our_NIPS} 			& \bf 68.2\% 	        &   AlexNet, AConv   \\
			 CrossLayer (proposed)        & \bf 68.2\%           &   AlexNet, AConv      \\ \\
			 CNN-Global				& 68.2\%          &   VGGNet               \\
			 CNN-Jitter				& 70.2\%            &   VGGNet                 \\
			 SCFV	\cite{Our_NIPS} 			& 73.5\% 	        &   VGGNet, OConv   \\
			 CrossLayer (proposed)        & \bf 74.4\%           &   VGGNet, OConv       \\
			 SCFV	\cite{Our_NIPS} 			& 76.4\% 	        &   VGGNet, AConv   \\
			 CrossLayer (proposed)        & \bf 78.2\%           &   VGGNet, AConv       \\
            \noalign{\smallskip}
            \hline
            \noalign{\smallskip}
            	 Fine-tuning \cite{Yao_IJCV}                   & 69.8\%      & fine-tunning with the VGGNet \\
             Fine-tuning \cite{ArxivNewBaseline}                         & 66.0\%      & fine-tunning with the AlexNet \\
			 MOP-CNN    \cite{CNN_Regional}         		&  68.9\%	  &  three scales   \\
             VLAD level2  \cite{CNN_Regional}  			& 65.5\%          &  single scale  \\
			 CNN-SVM      \cite{CNN_Baseline}            & 58.4\%          & - \\
			 FV+DMS   \cite{Dis_Mode_Seeking}    	& 63.2\%		     & -  \\
			 DPM               \cite{DPM}      & 37.6\%	         & - \\
			 DeepTexture \cite{Deep_texture} & \bf 81.7\% & 7 scales \\
			 Texture Synthesis \cite{Gatys2015b}  & 75.0\% & using the Gram matrix on fc18  \\
			                                      &      & layer (VGG net) for classification \\
            \hline
      \end{tabular}
    }
    \end{table}
We evaluate the proposed method on three datasets: MIT indoor scene-67 (MIT-67 in short) \cite{MIT67}, Caltech-UCSD Birds-200-2011 \cite{Birds200} (Birds-200 in short) and PASCAL VOC 2007 \cite{pascal-voc-2007} (PASCAL-07 in short) for image classification. These three datasets cover several popular topics in image classification, that is, scene classification, fine-grained object classification and generic object classification.

We compare the proposed method against three baselines, they are: (1) directly using fully-connected layer activations for the whole image (CNN-Global); (2) averaging fully-connected layer activations from several transformed versions of an input image. Following \cite{CNN_Baseline,ArxivNewBaseline}, we transform the input image by cropping its four corners and middle regions as well as by creating their mirrored versions (CNN-Jitter); (3) applying the sparse coding based Fisher vector encoding method \cite{Our_NIPS} on the local feature extracted from the convolutional layer (in both schemes) (SCFV). Since R-CNN SCFV has demonstrated superior performance to the MOP method in \cite{CNN_Regional}, we do not include MOP in our comparison. To make a fair comparison, we reimplement all three baseline methods. We also apply PCA, $\ell_2$ normalization and power normalization to SCFV and $\ell_2$ normalization to CNN-Global and CNN-Jitter (We find that using PCA and/or power normalization makes little difference to the result of CNN-Global and CNN-Jitter).

Two CNN models are adopted throughout our experiment: the Alex net \cite{ImageNetDeepLearning} and the VGG very-deep 19 layer network \cite{VGGVD}. Two different types of convolutional layers are used, the original convolutional layer (denoted as OConv) and the augmented convolutional layer (denoted as AConv). As discussed in section \ref{sect:fc_vs_conv}, the latter can be implemented by applying DCNN on a set of image regions which are extracted on a dense grid. In our implementation, we first resize the input image to 512$\times$512 pixels and then extract image regions with the size 128$\times$128 at a step size of 32 pixels. For OConv layers, we report the results obtained using the 4th and 5th convolutional layers for the Alex net and the conv5-3 and conv5-4 convolutional layers for the VGGVD net since those settings achieve the best performance. We also explore the use of other convolutional layers in the second part of our experiment. For the AConv layer, we extract local features from the fc6 layer in the Alex net and the first fully-connected layer in the VGGVD net respectively. Then we stack a new convolutional layer with 100 filters on top of them and train the newly added layer on the target dataset. This new layer is trained with the learning rate 0.01 by using 50 epochs. No data augmentation is used in this training step.

We use libsvm \cite{libsvm} as the SVM solver and use precomputed linear kernels as inputs. This is because the calculation of linear kernels/Gram matrices can be easily implemented in parallel. When  feature dimensionality is high, the kernel matrix computation actually occupies most of the computational time. Thus it is appropriate to use parallel computing to accelerate this process.

\subsubsection{Classification Results}

\noindent\textbf{Scene classification: MIT-67.}
MIT-67 is a commonly used benchmark for evaluating scene classification algorithms. It contains 6700 images with 67 indoor scene categories. Following the standard setting, we use 80 images in each category for training and 20 images for testing.  The results are shown in Table \ref{table:MIT67_Result}. It can be seen that the proposed cross-layer pooling achieves the overall best performance in most settings. The best performance is achieved by using cross-layer pooling and the AConv layer: this setting produces 68.2\% classification accuracy for the Alex net and 78.2\% for the VGGVD net. Also, it is clear that extracting local features from the AConv layer, as has been done in SCFV and CrossLayer, achieves significant performance increase in comparison with global CNN features, i.e. Global and Jitter. Finally, the use of the VGGVD net further boosts the classification performance by a large margin.

By comparing the performance reported from the literature, we can see that the proposed method surpasses most state-of-the-art methods. The only exception is the result in \cite{Deep_texture}. However, its method is very close to our SCFV (with OConv) baseline, and its good performance is largely due to their brute force multiple-scale strategy (they have utilized 7 scales while we only use a single scale).

\noindent\textbf{Fine-grained image classification: Birds-200.}
Birds-200 is the most popular dataset in fine-grained image classification research. It contains
11788 images with 200 different bird species. This dataset provides ground-truth annotations of
bounding boxes and parts of birds such as  the head and the tail, on both the training set and the test
set. In this experiment, we only use the bounding box annotation. The results are shown in Table \ref{table:Birds_result}. As can be seen, the cross-layer pooling achieves the best classification performance: 77.0\% when the VGGVD net is used. Also,
using the original convolutional layer achieves much better performance than the use of the AConv layers. For both the Alex net and the VGGVD net, the best performance is achieved by using features from the OConv layer. The underlying reason can be well explained by section \ref{sect:two_layer_types}, that is, the discriminative information of birds species usually lies at small regions and it will be more appropriate to extract features from original convolutional layers due to the image style mismatch issue discussed in section \ref{sect:two_layer_types}.

Our best performance is among the best for the dataset. The work in \cite{BilinearCNN} reports
higher classification accuracy than us. However, it relies on a fine-tuning step on two different
networks and it adopts some different experimental settings, e.g., their convolutional layers have a different number of spatial units to ours\footnote{When the same spatial units configuration is used, our cross-layer pooling  achieves 80\% classification accuracy which is closer to the result in \cite{BilinearCNN}. Note that we only use one network and do not apply fine-tuning and score calibration.} (28 $\times$ 28 as oppose to 14 $\times$ 14 in our experiments); it performs decision score calibration on the SVM while we just use the standard one-vs-the-rest SVM.

\begin{table}
        \caption{Comparison of results on Birds-200. Note that the method annotated with ``use parts'' requires parts annotations and detection while our methods do not employ these annotations so they are not directly comparable with our method. Also, the fine-tuning result in \cite{BilinearCNN} is achieved with a different configuration while our method achieves 80\% with the same configuration.    }
		\centering
		\label{table:Birds_result}
    	\scalebox{.952}
{
		\begin{tabular}{llll}
            \hline\noalign{\smallskip}
                Methods  &    Accuracy & Remark \\
            \noalign{\smallskip}
            \hline
            \noalign{\smallskip}
			 CNN-Global				& 59.2\%          &    no parts. AlexNet        \\
			 CNN-Jitter				& 60.5\%          &    no parts. AlexNet         \\
			 SCFV	\cite{Our_NIPS} 			& 64.2\% 	& no parts, AlexNet, OConv \\
			 CrossLayer         & \bf 73.3 \%          &    no parts, AlexNet, OConv \\
			 SCFV	\cite{Our_NIPS} 			& 66.4\% 	& no parts, AlexNet, AConv \\
			 CrossLayer         & 71.7\%           &    no parts, AlexNet, AConv \\
			 \\
			 CNN-Global				& 62.5\%          &    no parts. VGGNet        \\
			 CNN-Jitter				& 63.6\%          &    no parts. VGGNet         \\
			 SCFV	\cite{Our_NIPS} 			& 73.7\% 	&  no parts, VGGNet, OConv   \\
			 CrossLayer         & \bf 77.0\%           &    no parts, VGGNet, OConv \\
			 SCFV	\cite{Our_NIPS} 			& 66.2\% 	&  no parts, VGGNet, AConv   \\
			 CrossLayer         & 69.4\%           &    no parts, VGGNet, AConv \\
            \noalign{\smallskip}
            \hline
            \noalign{\smallskip}
                Fine-tuning \cite{BilinearCNN}                  &  76.4\%  & no parts, fine tunning, VGGNet \\
                Fine-tuning \cite{ArxivNewBaseline}                  &  66.4 \% & no parts, fine tunning, AlexNet \\
            	 Parts-RCNN-FT    \cite{ZhangNingECCV}                  & 76.37 \% & use parts, fine tunning \\
            	 Parts-RCNN    \cite{ZhangNingECCV}                  & 68.7 \% & use parts, no fine tunning \\
			 CNNaug-SVM    \cite{CNN_Baseline}         	& 61.8\%	  &  - \\
             CNN-SVM  \cite{CNN_Baseline}  			    & 53.3\%     & CNN global \\
			 DPD+CNN \cite{Decaffe}   & 65.0\%          & use parts \\
			 DPD   \cite{Zhang_2013_ICCV}    	            & 51.0\%		     & -  \\
			 Bilinear CNN   \cite{BilinearCNN}    	            & 77.9\%		     & Two networks  \\
			 Bilinear CNN   \cite{BilinearCNN}    	            & \bf 81.9\%		     & Two networks, fine-tuning  \\
			 Texture Synthesis \cite{Gatys2015b}  & 67.3 & using the Gram matrix on conv5-4  \\
			                                      &      & layer (VGG net) for classification \\
            \hline
      \end{tabular}
    }
\end{table}

\noindent\textbf{Object classification: PASCAL-2007.}
PASCAL VOC 2007 contains 9963 images with 20 object categories. The task is to predict the presence of each object in each image. Note that most object categories in PASCAL-2007 are also included in ImageNet which is the training set of the Alex net and the VGGVD net. So ImageNet can be seen as a superset of PASCAL-2007. The results on this dataset are shown in Table \ref{table:Pascal_result}. From Table \ref{table:Pascal_result}, we can see that again the best performance is achieved by using cross-layer pooling and the VGGVD net. Not surprisingly, the AConv layer performs better than the OConv layer in this dataset because the training categories of the DCNN overlaps with PASCAL-2007 and the AConv layer contains this category-level information. The per-class performance of three best comparing methods, that is, CNN jitter with the VGGVD net, SCFV with the AConv layer from the VGGVD net and cross-layer pooling with the AConv layer from the VGGVD net, is shown in Table \ref{table:Pascal07_Each_Class}. As seen, the proposed cross-layer pooling achieves the best performance in most classes.

\begin{table}
        \caption{Comparison of results on PASCAL VOC 2007. }
		\centering
		\label{table:Pascal_result}
	\scalebox{.952}
{
		\begin{tabular}{llll}
            \hline\noalign{\smallskip}
                Methods  &   mAP & Remark  \\
            \noalign{\smallskip}
            \hline
            \noalign{\smallskip}
			 CNN-Global				& 71.7\%          &   AlexNet  \\
			 CNN-Jitter				& 75.0\%          &   AlexNet \\
			 SCFV	\cite{Our_NIPS} 			& 66.8\% 	&   AlexNet, OConv   \\
			 CrossLayer         & 71.3\%           &   AlexNet, OConv      \\
			 SCFV	\cite{Our_NIPS} 			& 76.9\% 	&   AlexNet, AConv   \\
			 CrossLayer         & \bf 79.1\%           &   AlexNet, AConv     \\
			 \\
			 CNN-Global				& 83.6\%          &   VGGNet  \\
			 CNN-Jitter				& 84.5\%          &   VGGNet \\
			 SCFV	\cite{Our_NIPS} 			& 82.9\% 	&   VGGNet, OConv   \\
			 CrossLayer         & 84.1\%           &   VGGNet, OConv            \\
			 SCFV	\cite{Our_NIPS} 			& 85.1\% 	&   VGGNet, AConv   \\
			 CrossLayer         & \bf 87.1\%           &   VGGNet, AConv            \\
            \noalign{\smallskip}
            \hline
            \noalign{\smallskip}
             Fine-tuning \cite{RegionalGating}			&		\bf 90.1\% 	& VGGNet	 fine-tuning	\\
             Fine-tuning \cite{ReturnDevil}			&		82.4\%	& CNN-S fine tuning		\\
			 CNNaug-SVM    \cite{CNN_Baseline}         	&  77.2\%	  & with augmented data  \\
             CNN-SVM      \cite{CNN_Baseline}  			& 73.9\%      & no augmented data  \\
			 NUS   \cite{NUS}    				& 70.5\%		     & -  \\
			 GHM   \cite{GHM}                   & 64.7\%	         & - \\
			 AGS  \cite{AGS}            		& 71.1\%           & - \\
			 Texture Synthesis \cite{Gatys2015b}  & 84.7\% & using the Gram matrix on fc18  \\
			                                      &      & layer (VGG net) for classification \\
            \hline
      \end{tabular}
    }
\end{table}

\begin{table*}[!ht]
        \caption{Comparison of results on Pascal VOC 2007 for each of 20 classes. The classes on which cross-layer pooling achieves significantly better performance are labeled with bold font.}
        	\centering
		\label{table:Pascal07_Each_Class}
		\begin{tabular}{llllllllllll}
		\hline\noalign{\smallskip}
  & TV & train & sofa & sheep & plant & person & mbike & horse & dog & table \\
         \noalign{\smallskip}
         \hline
         \noalign{\smallskip}
Global Jitter (VGGVD) &  81.9 & 96.3 & 72.9 & 86.1 & 61.6 & 95.2 & 89.3 & 91.5 & 90.9 & 79.7 \\
SCFV (VGGVD) & 82.3 & 95.7 & 78.9 & 84.0 & 62.6 & 95.8 & 89.1 & 94.1 & 90.4 & 82.2 \\
CrossLayer (VGGVD) & \bf 85.3 & 96.7 &  \bf 80.1 & \bf 87.6 & \bf 64.6 & \bf 96.7 & \bf 91.7 & 94.3 & \bf 93.0 & 82.1 \\
         \noalign{\smallskip}
         \hline
         \noalign{\smallskip}
  &cow & chair & cat & car & bus & bottle & boat & bird & bike & areo \\
           \noalign{\smallskip}
         \hline
         \noalign{\smallskip}
Global Jitter (VGGVD) & 77.8 & 67.4 & 92.1 & 91.2 & 85.2 & 56.6 & 92.8 & 92.9 & 90.4 & 97.1 \\
SCFV (VGGVD) & 79.8 & 66.5 & 92.4 & 91.5 & 86.1 & 59.6 & 90.7 & 93.2 & 90.9 & 96.6 \\
CrossLayer (VGGVD) &  \bf 83.3 & \bf 70.2 & \bf 93.9 & \bf 92.8 & \bf 89.4 & 59.0 & \bf 93.8 & \bf 94.7 & \bf 94.0 & 97.8 \\
\hline
      \end{tabular}
\end{table*}

\subsection{Ablation study}
From the above experiments, the advantage of using the proposed method has been clearly demonstrated. In this section, we further examine the effect of various components in our method.

\subsubsection{Using different convolutional layers}\label{sect: different layers}
First, we are interested in examining the performance of using convolutional layers other than the 4th and 5th convolutional layers in the Alex net and the conv5-2 and conv5-3 convolutional layers in the VGGVD net. We investigate the performance of using the 3rd and 4th convolutional layers for the Alex net and the conv5-2 and conv5-3 convolutional layers in the VGGVD net. The results are shown in Table \ref{table:3-4 Layer result}. From the results, we can see that using 4-5th layers (conv5-3-4th) layers achieves superior performance over using the 3-4th layers (conv5-2-3th) layers. This is consistent with the observation in~\cite{VisualizeCNN} that the deeper the convolutional layer, the better discriminative power it has.

As discussed in section \ref{sect:fc_vs_conv}, the process of extracting fully-connected layer
activations from multiple local regions can be viewed as applying a special convolutional layer.
Thus it is possible to perform cross-layer pooling on two fully-connected layers. For AConv layers, a
new fully-connected layer is stacked and re-trained. Certainly, it is also possible to directly use
two fully-connected layers in an existing CNN without introducing new layers, but the
computational cost can be higher due to the high-dimensionality of the resulting representation,
e.g.,  4096$\times$1000. Also, it is possible to perform cross-layer pooling on an original convolutional layer and a fully-connected layer in the above setting. In such cases, multiple spatial units in a convolutional layer will correspond to one fully-connected layer output, to apply cross-layer pooling we can either flatten activations from multiple spatial units into a long vector or using the pooled activation from multiple spatial units. In this paper, we use the latter approach since it produces lower dimensional image representations.

In this subsection we conduct an experimental evaluation of the above two approaches to performing cross-layer pooling. Specifically, for the first approach, denoted as FC-FC cross-layer pooling, we use the activations from the first fully-connected layer (4096 dimensions) and the last fully-connected layer (1000 dimensions) in the VGG net to perform cross-layer pooling. PCA is applied to reduce the dimensionality of the first fully-connected layer activations to 2000; for the second approach, denoted as FC-Conv cross-layer pooling, we use the activations from conv5-4 and the first fully-connected layer in VGG net and apply sum-pooling (followed by using square-root post-processing) to pool the activations of conv5-4. Thus the dimensionality of the representation obtained from cross-layer pooling will be 512$\times$4096. The performance of these two methods is shown in Table \ref{table:fc Layer result}. From Table \ref{table:fc Layer result} we make the following two observations: (1) by cross-referencing the performance in Table \ref{table:MIT67_Result}, Table \ref{table:Birds_result} and Table \ref{table:Pascal_result}, we  observe that FC-FC cross-layer pooling achieves similar performance to cross-layer pooling using the AConv layer,
but marginally worse.
This may suggest that the good performance of the AConv layer based cross-layer pooling mainly comes from the cross-layer pooling strategy, although the re-trained new convolutional layer can further boost classification performance. (2) FC-Conv cross-layer pooling achieves better performance than FC-FC cross-layer pooling on Birds200 but it is inferior to FC-FC cross-layer pooling on MIT-67 and PASCAL-07. Considering that the pooling operation of FC-Conv cross-layer pooling is performed on the OConv layer, this observation is consistent with the conclusion in Table \ref{table:MIT67_Result} and Table \ref{table:Pascal_result}, that is, cross-layer pooling on AConv layers leads to better performance than pooling with OConv layers for MIT-67 and PASCAL-07.

\begin{table}
        \caption{Comparison of results obtained by using cross-layer pooling with fully-connected layers.}
		\centering
		\label{table:fc Layer result}
    	\scalebox{1.0}
{
		\begin{tabular}{llllll}
            \hline\noalign{\smallskip}
                Method  &   MIT-67 & Birds200 & PASCAL07  \\
            \noalign{\smallskip}
            \hline
            		FC-18, FC-20 (VGGNet)    &  77.0\%  &  63.2\%     &  85.9\%  \\
            		conv5-3, FC-18 (VGGNet)  &  74.1\%  &  68.5\%  &  84.5\%  \\
            \noalign{\smallskip}
            \hline
      \end{tabular}
    }
\end{table}

\begin{table}
        \caption{Comparison of results obtained by using different convolutional layers.}
		\centering
		\label{table:3-4 Layer result}
    	\scalebox{.952}
{
		\begin{tabular}{llllll}
            \hline\noalign{\smallskip}
                Method  &   MIT-67 & Birds200 & PASCAL07  \\
            \noalign{\smallskip}
            \hline
            		CL-3-4 (AlexNet)    &   59.4\% & 63.9\%  & 66.3\%  \\
            		CL-4-5 (AlexNet)      & \bf 63.0\%   & \bf 73.5\%  & \bf 71.3\%  \\
            		CL-conv5-2-3 (VGGNet)      &  73.7\%   & \bf 77.0\%  & 82.2\%  \\
            		CL-conv5-3-4 (VGGNet)      & \bf 74.4\%   & \bf 77.0\%  & \bf 84.1\%  \\
            \noalign{\smallskip}
            \hline
      \end{tabular}
    }
\end{table}

\subsubsection{The impact of PCA and normalization}
In our implementation, we have applied three operations to obtain the final image representation, that is,
performing PCA on the local feature, performing $\ell_2$ normalization on each pooled coding vector and power normalization. In this section, we investigate the impact of those three operations. We conduct our experiment on MIT-67 with the AConv layer features and test the performance under various settings of those three operations. Table \ref{table:implementation_comparison} shows the results. From Table \ref{table:implementation_comparison}, we observe some interesting phenomena: (1) The three operations have a big impact on the performance. If none of them is applied, the performance drops significantly. (2) Applying either $\ell_2$ normalization or power normalization leads to similar performance improvement.
(3) Applying PCA with normalization, $\ell_2$ normalization or power normalization or both of them, can lead to further performance improvement. (4) The best performance is obtained by applying all three operations together.

\begin{table}
        \caption{The impact of PCA, $\ell_2$ normalization and power normalization.}
		\centering
		\label{table:implementation_comparison}
    	\scalebox{.952}
{
		\begin{tabular}{ccccc}
            \hline\noalign{\smallskip}
                PCA  &   $\ell_2$ normalization & power normalization & Result  \\
            \noalign{\smallskip}
            \hline
            		Yes & Yes & Yes & \bf 78.2\%  \\
            		Yes & Yes &  -    & 76.4\%  \\
            		Yes & -  & Yes & 77.1\%  \\
            		- & Yes & Yes & 72.6\%  \\
            		Yes & -  & - & 69.7\%  \\
            		 - & Yes &  - & 74.6\%  \\
            		 -  &  - & Yes & 73.2\%  \\
            		 -    &  - &  - & 69.7\%  \\
            \noalign{\smallskip}
            \hline
      \end{tabular}
    }
\end{table}

\subsubsection{Feature sign quantization}\label{sect:coarse_quantization}
As has been discussed in Section \ref{sect:implementation_details}, feature sign quantization is a promising strategy to reduce the memory cost of cross-layer pooling. Here we demonstrate the effect of applying feature sign quantization. Feature sign quantization quantizes a feature to 1 if it is positive, -1 if it is negative and 0 if it equals  0. In other words, we use 2 bits to represent each dimension of the pooled feature vector. This scheme greatly reduces the memory use. Here we only report the result on the best performing setting for each dataset. The results are shown in Table \ref{table:FS_Quantization}. As can be seen, this coarse quantization scheme does not degrade the performance much, and for two datasets, MIT-67 and Birds-200, it achieves almost the same performance as the original feature. Note that a similar quantization scheme has been also explored in \cite{ExistingConvEx}, however there is caused a significant performance drop if applied to convolutional layer features. For example, in the Table 7 of \cite{ExistingConvEx}, by binarizing conv-5, the performance drops around 5\%. In contrast, our representation seems to be less sensitive to this coarse quantization.

\subsection{Comparison with alternative pooling methods}
Finally, we compare several alternative pooling strategies against cross-layer pooling.
The baseline method that we compare against directly pools convolutional layers.
There are many possible variants of such an approach, however, which we characterize according to three criteria:
 \begin{itemize}
   \item
Pooling methods. We consider both sum-pooling with square-root post-processing (which is better than direct sum-pooling) and max-pooling.
\item
  Spatial pyramids \cite{SPM}. Three spatial pyramid partitions, that is, the $1\times1$ (level 0), $1\times1 + 2\times2$ (level 1) and $1\times1+2\times2+4\times4$ (level3) are considered.
\item
  Pooling layers. The conv5-3 layer, conv5-4 layer and the concatenation of conv5-3 and conv5-4 layers of the VGG net are considered.
  \end{itemize}
  The classification results on MIT67, Birds200 and PASCAL07 are reported in Table \ref{table:conv_pooling_big_comparison}. As can be seen, the best performance achieved by tuning those pooling strategies is 69.0\% on MIT67, 64.1\% on Birds200 and 77.4\% on PASCAL07. In comparison, the cross-layer pooling counterpart achieves 74.4\%, 77.0\% and 84.1\% on MIT67, Birds200 and PASCAL07 respectively, which is much better than the traditional pooling approaches.

Another possible variation of cross-layer pooling is to change the channel pooling method from sum-pooling to max-pooling. We also tried this setting and achieved 72.8\% on MIT67, 71.6\% on Birds200 and 85.2\% on PASCAL07. As can be seen, its performance is worse than sum-pooling. Thus we suggest using sum-pooling as the default pooling method for cross-layer pooling.

%

\subsection{Experiments for image retrieval}\label{sect:retrieval experiment}
\begin{figure*}[!ht]
    \centering
    \begin{tabular}{l}
            \subfloat[]{ \includegraphics[height=45mm]{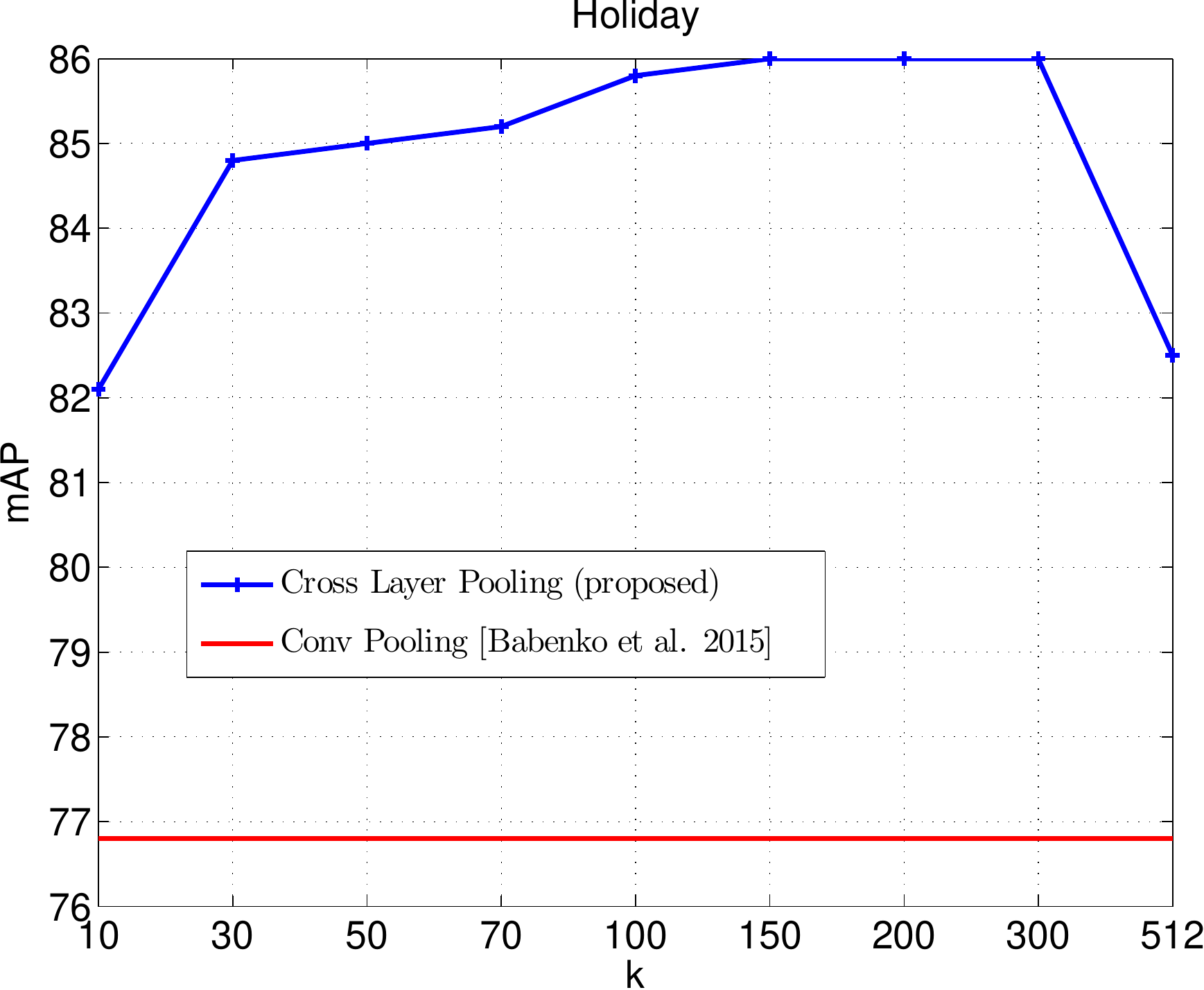}}
            \subfloat[]{ \includegraphics[height=45mm]{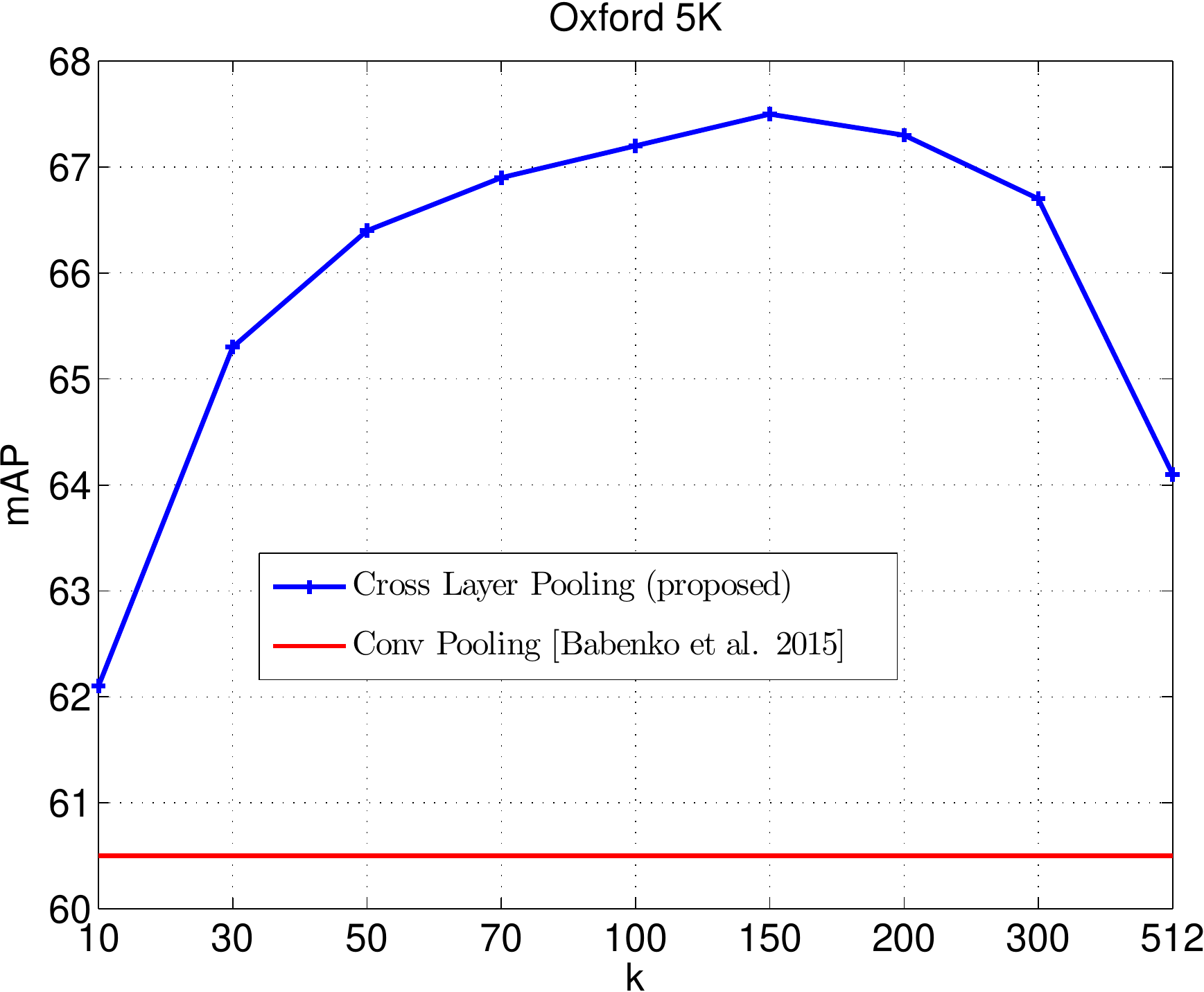}}
            \subfloat[]{ \includegraphics[height=45mm]{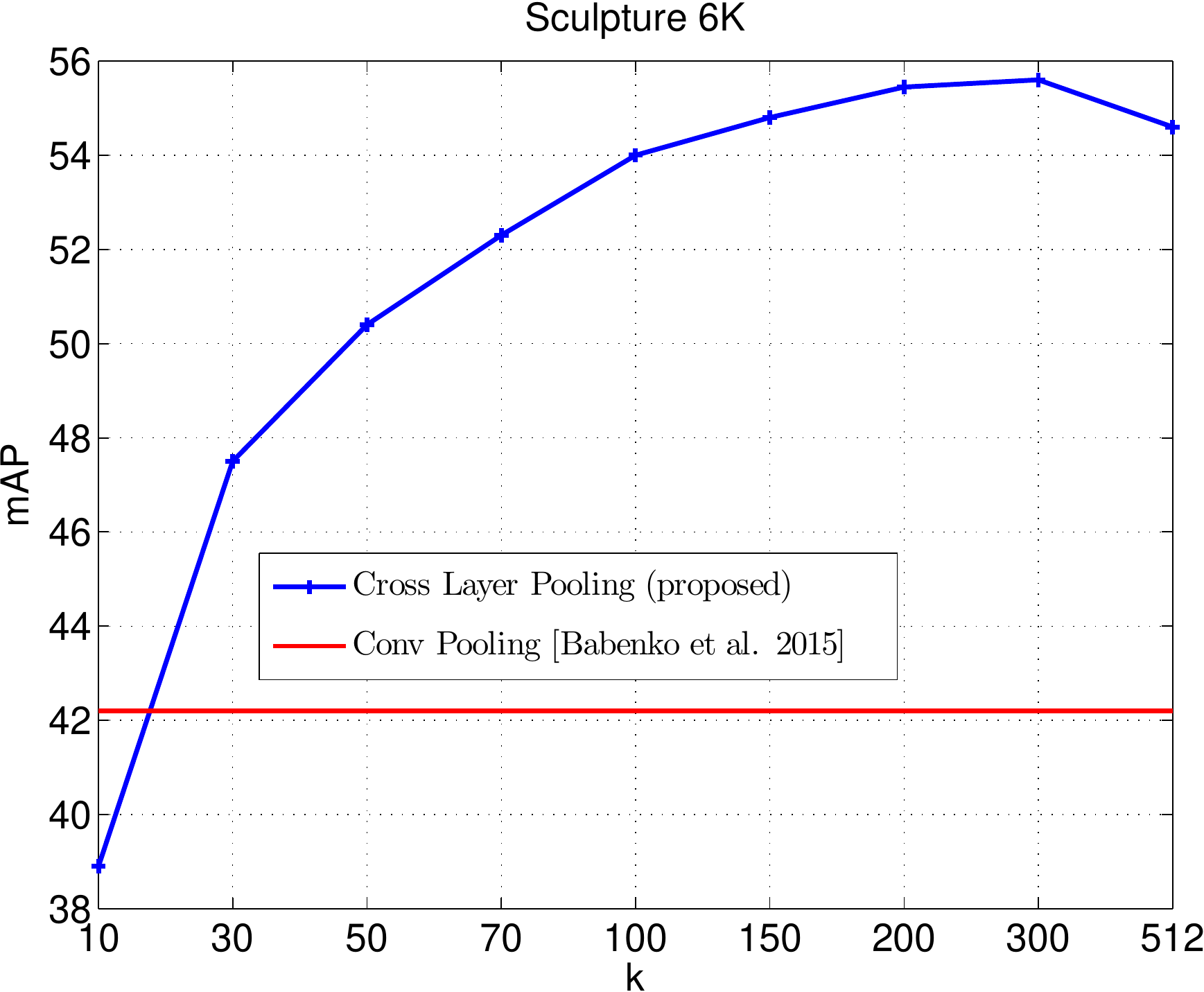}}
    \end{tabular}
    \caption{Performance of cross-layer pooling on image retrieval.
      For our method, the feature maps of the conv5-4 layer with top $k$ average activations are selected as pooling channels.
    }
    \label{fig:Retrieval_CL}
    \end{figure*}

For image retrieval, we evaluate cross-layer pooling on the Oxford5K \cite{Philbin07}, Holiday
\cite{Holiday_dataset} and Sculpture6K \cite{Arandjelovic11} datasets. These three datasets
represent several common scenarios in image retrieval. The objects-of-interest in Oxford5K are
buildings which have rich texture patterns. For the Holiday dataset, the to-be-retrieved images are more
general scenes and objects. The sculpture6K dataset focuses on sculptures which have relatively
smooth surfaces. We adopt a similar setting to~\cite{conv_pool_retrieval} to evaluate performance by
directly using the query image without the object bounding box. We re-implemented the baseline method
of \cite{conv_pool_retrieval} by strictly following its experimental protocol. Specifically, we
apply PCA whitening and calculate the PCA projection matrix from external datasets. For Oxford5K, we
learn the PCA projection matrix from Paris6K; for Holiday, we learn the PCA projection matrix from
5K Flickr images (although those 5K Flickr images might not be same as the one used in
\cite{conv_pool_retrieval}); for Sculpture, we also calculate the PCA projection matrix from the
same 5K Flickr images. The VGG net is used in this experiment, the baseline in
\cite{conv_pool_retrieval} is performed on the conv5-4 layer since it leads to the best performance.
Our approach is applied on conv5-3 and conv5-4 layers. We use binarized cross-layer pooling vectors
and select the pooling channels with top-$k$ largest average activation value on conv5-4. Different
$k$ are evaluated and the comparison of cross-layer pooling and the baseline method in
\cite{conv_pool_retrieval} is shown in Figure \ref{fig:Retrieval_CL}. From Figure
\ref{fig:Retrieval_CL}, it is clear that cross-layer pooling performs much better than the direct
convolutional layer pooling baseline \cite{conv_pool_retrieval} once a sufficiently large $k$ is
chosen.
Selecting $k=50$ is typically sufficient to achieve good performance.
Considering that all
the features are binarized, the computational cost is still reasonable
despite the fact that the dimensionality is higher than that of
the baseline method. Note that if we directly binarize the feature obtained
from the convolutional layer pooling baseline, this leads to significant performance drop, while our method
does not. Also, it is interesting to discover that when $k$ becomes too large,
that is, when we are close to using all of the available
pooling channels, the retrieval performance will start to drop. This is probably because including channels with small average activation values tend to introduce more noise  during retrieval.

\begin{table*}[ht!]
        \caption{Comparison of alternative pooling methods. Two pooling methods, max-pooling and sum-pooling (with square-root post-processing) are applied. Different levels of spatial pyramid \cite{SPM} configuration are applied for both pooling methods. Level 0: pooling over the whole image (1$\times$1); level 1: 1$\times$1+2$\times$2; level 2: 1$\times$1+2$\times$2+4$\times$4; Those pooling methods are applied to the conv5-3 layer, the conv5-4 layer and the concatenation of conv5-3 and conv5-4 layers of the VGG network. The experimental comparison is made on MIT67, Birds200, and PASCAL07 three datasets. Results are reported in the order MIT67/Birds200/PASCAL07. For reference, cross-layer pooling on conv5-3 and conv5-4 achieves classification accuracy 74.4/77.0/84.1, which is higher than the results of all the alternative pooling methods.}
		\centering
		\label{table:conv_pooling_big_comparison}
    	\scalebox{1.0}
{
		\begin{tabular}{llllll}
            \hline\noalign{\smallskip}
                Method  &   Conv5-3 & Conv5-4 & Conv5-3 + Conv5-4  \\
            \hline\noalign{\smallskip}
      			max-pooling, SPM level0   &  57.3/67.7/75.5  &  60.7/69.0/80.5 & 60.6/55.8/77.8 \\
				max-pooling, SPM level1   &  66.3/67.6/76.8  &  67.7/70.2/81.6 & 66.0/59.1/78.9 \\
				max-pooling, SPM level2   &  68.0/67.6/76.8  &  69.3/69.9/81.9 & 67.9/61.5/79.8    \\
		  sum-sqrt-pooling, SPM level0    &  66.4/66.4/77.7  &  68.2/69.3/81.7 & 69.9/65.8/80.9 \\
		  sum-sqrt-pooling, SPM level1    &  68.0/64.1/77.5  &  70.0/69.8/81.7 & 70.7/63.6/80.6 \\
		  sum-sqrt-pooling, SPM level2    &  69.0/64.1/77.4  &  70.3/69.8/81.9 & 70.9/64.1/80.7  \\
            \noalign{\smallskip}
            \hline
      \end{tabular}
    }
\end{table*}

\begin{table}
        \caption{Results obtained by using feature sign quantization.}
		\centering
		\label{table:FS_Quantization}
		\begin{tabular}{lcc}
            \hline\noalign{\smallskip}
                Dataset  &   Feature sign quantization & Original \\
            \noalign{\smallskip}
            \hline
            		MIT-67 (VGG, Aconv)           &  77.9\% & 78.2\% \\
            		Birds-200 (VGG, OConv)        &  76.5\%  & 77.0\% \\
            		PASCAL07 (VGG, AConv)         &  85.1\% & 87.0\% \\
            \noalign{\smallskip}
            \hline
      \end{tabular}
\end{table}

\section{Conclusion}
We have proposed a new method termed cross-convolutional layer pooling to create image representations from the activations of two consecutive convolutional layers of a pre-trained CNN. We realize this idea on two types of implementations of convolutional layers and show that these two different implementations are particularly well suited to different recognition tasks. Also, we propose a variation on the cross-convolutional layer pooling approach for the image retrieval task. By conducting experiments on popular image classification datasets and image retrieval datasets, we show that the proposed method leads to superior performance over various existing methods of using a pre-trained DCNNs to extract image representations.

\section*{Acknowledgments}

This work was in part supported by
the Data to Decisions Cooperative Research Centre; and Australian Research Council Future Fellowship
(FT120100969), and Australian Research Council projects DP160103710, and LP130100156.

C. Shen is the corresponding author.

\IEEEpeerreviewmaketitle

\bibliographystyle{ieee}
\bibliography{CSRef_2}

\end{document}